\documentclass[10pt,twocolumn,letterpaper]{article}

\usepackage[pagenumbers]{cvpr}
\usepackage{float}
\usepackage[colaction]{multicol}
\usepackage{needspace}
\usepackage{colortbl}
\definecolor{modelgroupblue}{RGB}{225,239,250}

\usepackage{microtype}

\definecolor{cvprblue}{rgb}{0.21,0.49,0.74}
\usepackage[pagebackref,breaklinks,colorlinks,allcolors=cvprblue]{hyperref}

\title{PathStream: Towards Streaming Diagnostic Copilots for Whole-Slide Pathology}
\author{%
  Dankai Liao$^{1,\ddagger}$ \quad YuFeng Wu$^{2,\ddagger}$ \quad Ziyue Wang$^{1,\ddagger}$ \quad Guanxi Liu$^{2}$\\
  Tianyi Zhang$^{1,2}$ \quad Zeyu Liu$^{2}$ \quad Linghan Cai$^{2,3,*}$ \quad Yueming Jin$^{1,\dagger}$\\[4pt]
  {\normalsize $^{1}$National University of Singapore \quad $^{2}$PuzzleLogic Pte. Ltd. \quad $^{3}$Heidelberg University}
}
\hypersetup{
  pdftitle={\thetitle},
  pdfauthor={Dankai Liao, YuFeng Wu, Ziyue Wang, Guanxi Liu, Tianyi Zhang, Zeyu Liu, Linghan Cai, Yueming Jin}
}

\begin{document}
%\raggedbottom
\maketitle
\begingroup
  \renewcommand{\thefootnote}{\fnsymbol{footnote}}
  \footnotetext[3]{Equal contribution.}
  \footnotetext[1]{Corresponding author\enspace \href{mailto:linghancai@puzzlelogic.com}{\nolinkurl{linghancai@puzzlelogic.com}}}
  \footnotetext[2]{Lead corresponding author\enspace \href{mailto:ymjin@nus.edu.sg}{\nolinkurl{ymjin@nus.edu.sg}}}
\endgroup

\begin{abstract}
Pathologists reach a diagnosis by integrating morphological evidence across regions and magnifications along viewing trajectories. Existing pathology AI rarely reflects this process: models predict from selected image regions, answer questions within fixed visual contexts, or navigate slides autonomously. We formalize \emph{pathologist-led streaming diagnosis}, in which the pathologist controls navigation while a copilot selectively describes incoming views and synthesizes accumulated evidence into a final diagnosis. We record five pathologists' viewing trajectories on 1,791 TCGA whole-slide images (WSIs) spanning ten organs, with response labels, morphological descriptions, and session-level findings and diagnoses. We construct \textbf{PathStreamBench} to assess response timing, morphological description, and session-level diagnosis, alongside auxiliary video understanding. We introduce \textbf{PathStream}, which jointly learns streaming responses and diagnosis through two-stage training, with a diagnosis-attention loss guiding attention during diagnosis to relevant visual and textual evidence accumulated as the examination unfolds. PathStream outperforms 14 vision-language models and three domain-fine-tuned baselines on the core tasks, achieving a trajectory WSI Relevance score of 77.87, surpassing the strongest domain-fine-tuned baseline's 62.76. Without additional training, its diagnostic capabilities generalize to new educational pathology videos, three external centers, and an unseen organ.
\end{abstract}
\section{Introduction}
\label{sec:introduction}

\begin{figure}[t]
    \centering
    \includegraphics[width=\linewidth]{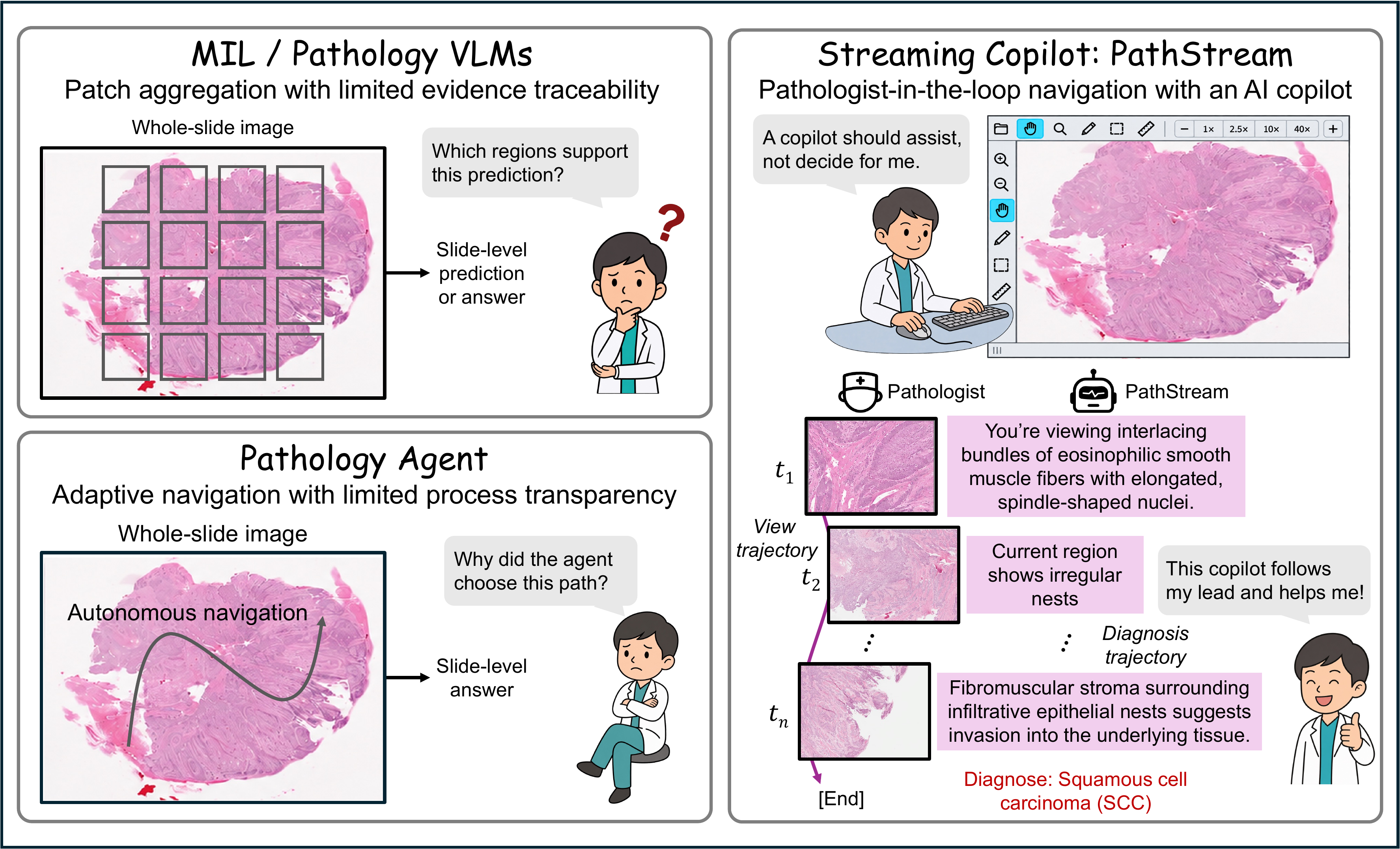}
    \caption{\textbf{Motivation of streaming diagnostic copilots.}
    Static models such as multiple instance learning (MIL) and pathology vision-language models (VLMs) aggregate patches or answer questions about a given visual context, while pathology agents select their own viewing  trajectories. The proposed PathStream follows the pathologist's navigation, describes the evolving morphology, and integrates accumulated evidence into a  final diagnosis.}
    \label{fig:introduction}
\end{figure}

Histopathological diagnosis develops through a sequence of observations: pathologists examine different tissue regions of a whole-slide image (WSI) at multiple magnifications, interpret local morphology, and combine findings to reach a diagnostic conclusion. These viewing trajectories vary with individual experience and clinical practice. However, existing pathology AI is generally not designed for this \emph{ongoing, pathologist-led examination}: static approaches make slide-level predictions from a selection of image regions or answer questions based on fixed visual context~\cite{luDataefficientWeaklySupervised2021a,shaoTransMILTransformerBased2021,luMultimodalGenerativeAI2024a,seyfiogluQuiltLLaVAVisualInstruction2024}, while autonomous agents select their own trajectories~\cite{sunCPathAgentAgentbasedFoundation2025a,wangPathologyCoTLearningVisual2026a}. We therefore formalize \emph{pathologist-led streaming diagnosis} (\cref{fig:introduction}), in which the pathologist directs slide navigation while a diagnostic copilot follows the evolving examination to interpret findings and support diagnosis. By preserving the pathologist's control over navigation, this setting is designed to accommodate diverse examination practices and integrate AI assistance into existing diagnostic workflows.

An ideal streaming copilot in this setting should therefore adapt its responses as new views arrive, providing timely descriptions of newly observed morphology while remaining silent during continued inspection of similar views. Its descriptions should accurately capture the morphology observed in different tissue regions at multiple magnifications. Beyond interpreting individual views, the copilot should retain and integrate relevant findings from earlier observations to produce a final diagnosis at session end.

Training such a copilot requires annotated viewing trajectories that link response timing and morphological interpretation with diagnostic conclusions. We therefore record viewing trajectories as five pathologists examine 1,791 WSIs from 1,791 patients across ten organs in The Cancer Genome Atlas (TCGA)~\cite{thecancergenomeatlasresearchnetworkCancerGenomeAtlas2013a}. Each trajectory is paired with response labels, morphological descriptions for response frames, and session-level microscopic findings and diagnoses. From the held-out test video sessions, we construct \textbf{PathStreamBench} with four tasks: response timing, streaming morphological description, session-level diagnosis, and auxiliary video understanding covering visual content and temporal relations. External validation assesses diagnostic generalization beyond TCGA: a private three-center cohort tests transfer to new clinical centers, while the PEAN skin cohort~\cite{nanDeepLearningQuantifies2025} tests generalization to unseen organ.

We further introduce \textbf{PathStream}, a streaming copilot for \emph{pathologist-led diagnosis} that follows the pathologist's slide navigation, responds selectively as new views arrive, and integrates accumulated findings into a final diagnosis. We train PathStream in two stages: the first establishes basic streaming capabilities in response control, morphological description, and diagnosis; the second focuses on strengthening session-level diagnostic synthesis. Across both stages, a diagnosis-attention loss guides the model to integrate accumulated frames and descriptions during diagnosis generation. On PathStreamBench, we compare PathStream with 14 general-purpose, streaming, and pathology/medical vision-language models (VLMs) and fine-tuned baselines. PathStream outperforms all baselines in the response timing, morphological description, and diagnostic synthesis required for streaming diagnosis. Without additional training, its diagnostic capabilities also generalize to educational pathology videos, new clinical centers, and an unseen organ. In summary, the contributions of this work are:
\begin{itemize}
    \item We formalize \emph{pathologist-led streaming diagnosis}: the pathologist controls slide navigation while a copilot selectively describes morphology as new views arrive and integrates accumulated evidence into a final diagnosis.
    \item We curate viewing trajectories from 1,791 TCGA WSIs spanning ten organs, annotated for response timing, morphological description, and session-level diagnosis. We construct \textbf{PathStreamBench} to evaluate these capabilities, complemented by an auxiliary video understanding task.
    \item We introduce \textbf{PathStream}, which jointly learns streaming responses and diagnostic synthesis and outperforms both original and domain-fine-tuned VLMs on the core streaming diagnostic tasks. It achieves a trajectory WSI Relevance score of 77.87, compared with 62.76 for the strongest domain-finetuned baseline, and generalizes without additional training to educational pathology videos, 3 new clinical centers, and an unseen organ.
\end{itemize}

\section{Related Work}
\label{sec:related_work}

\noindent \textbf{Whole-slide Modeling in Pathology.}
In whole-slide prediction, multiple-instance learning (MIL) aggregates
patch features under slide-level supervision. Attention pooling
and clustering-based objectives identify informative tissue
regions~\cite{ilseAttentionbasedDeepMultiple2018,luDataefficientWeaklySupervised2021a}.
Transformers and hierarchical encoders model relationships between regions
and across scales~\cite{shaoTransMILTransformerBased2021,
chenScalingVisionTransformers2022}, while large-scale pretraining produces
transferable slide representations~\cite{xuWholeslideFoundationModel2024}.
A complementary direction learns from how pathologists examine slides:
eye-tracking records guide regional attention, and viewport trajectories
support prediction of viewing
paths~\cite{nanDeepLearningQuantifies2025,chakrabortyMeasuringPredictingWhere2026}.

\noindent \textbf{Pathology Vision-Language Models and Agents.}
Pathology VLMs align images with descriptions and reports to support
recognition and
retrieval~\cite{luVisuallanguageFoundationModel2024b,dingMultimodalWholeslideFoundation2025}.
Instruction tuning extends these representations to morphological
description and diagnostic question
answering~\cite{luMultimodalGenerativeAI2024a,seyfiogluQuiltLLaVAVisualInstruction2024},
with models operating on whole
slides~\cite{liangWSILLaVAMultimodalLarge2025} or pathology video
clips~\cite{vuongVideoPathLLaVAPathologyDiagnostic2025}.
Reasoning supervision and reinforcement learning further support
diagnostic reasoning~\cite{zhangPathoR1MultimodalReinforcement2026b}.
VLM-based agents select regions and change magnification to collect
diagnostic
evidence~\cite{sunCPathAgentAgentbasedFoundation2025a}.
Recorded expert navigation, gaze, and narration can guide their visual
grounding and examination
procedures~\cite{wangPathologyCoTLearningVisual2026a,wangPathCoworkLearningRealworld2026}.
These agents gather evidence through model-directed navigation.
PathStream instead follows the pathologist's examination,
jointly learning response timing, morphological interpretation,
and diagnostic synthesis.

\begin{figure*}[tp]
    \centering
    \includegraphics[width=\linewidth]{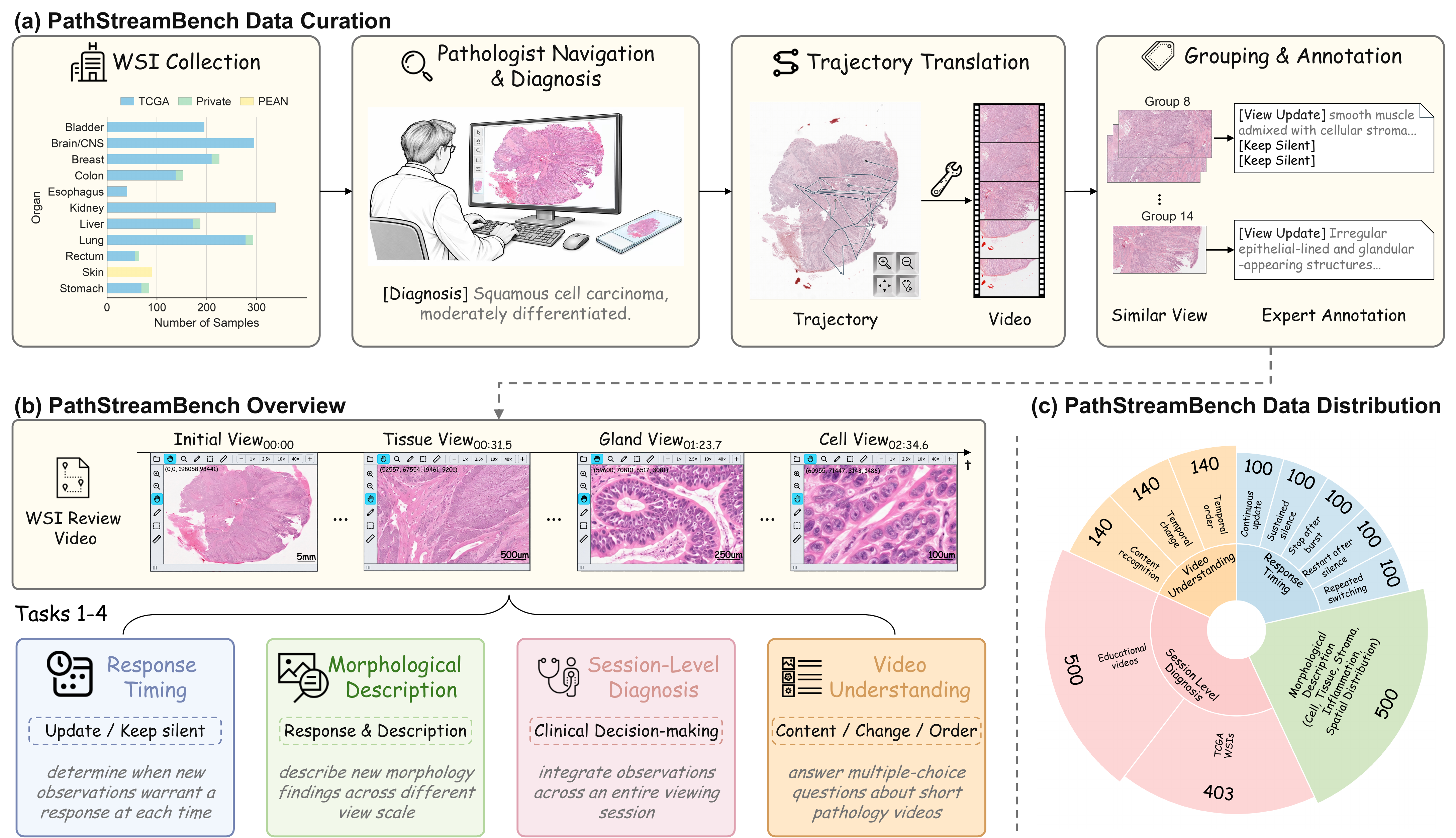}
    \caption{\textbf{Data curation and PathStreamBench construction.}
    (a) Organ distribution across the TCGA, private, and PEAN cohorts,
    followed by pathologist-guided slide review, trajectory replay,
    and expert annotation of response points and morphology.
    (b) The benchmark evaluates response timing, streaming morphological
    description, session-level diagnosis, and auxiliary pathology video
    understanding along the viewing sequence.
    (c) Distribution of PathStreamBench samples across the four tasks
    and their categories.}
    \label{fig:benchmark_overview}
\end{figure*}

\noindent \textbf{Streaming Video Understanding.}
Streaming video understanding studies how models interpret and respond to
visual inputs as they arrive. Temporally aligned training interleaves
frames and language~\cite{chenVideoLLMonlineOnlineVideo2024a,
chenLiveCCLearningVideo2025}.
Recent methods learn when to respond through explicit decision modules
or reinforcement learning~\cite{qianDispiderEnablingVideo2025a,
wangMMDuet2EnhancingProactive2026}, retain historical context through memory
or cached states~\cite{zhangFlashVstreamEfficientRealTime2025,
xuStreamingVLMRealTimeUnderstanding2026a}, and reason while continuing to
observe~\cite{guanVideoStreamingThinking2026}.
In parallel, benchmarks evaluate current-event understanding and contextual
reasoning~\cite{linStreamingbenchAssessingGap2026}, recall of earlier events
and responses triggered by new evidence~\cite{niuOVOBenchHowFar2025}, and
sustained frame--text alignment over long
streams~\cite{xuStreamingVLMRealTimeUnderstanding2026a}.
PathStreamBench extends streaming evaluation to pathologist-led streaming diagnosis
through four tasks: response timing, streaming morphological description,
session-level diagnosis, and auxiliary pathology video understanding.
These tasks assess timely interpretation of tissue observations
and diagnostic synthesis from accumulated evidence.
\section{Pathologist-Led Streaming Diagnosis}
\label{sec:Pathologist-Led Streaming Diagnosis}

\begin{figure*}[t]
    \centering
    \includegraphics[width=0.92\linewidth]{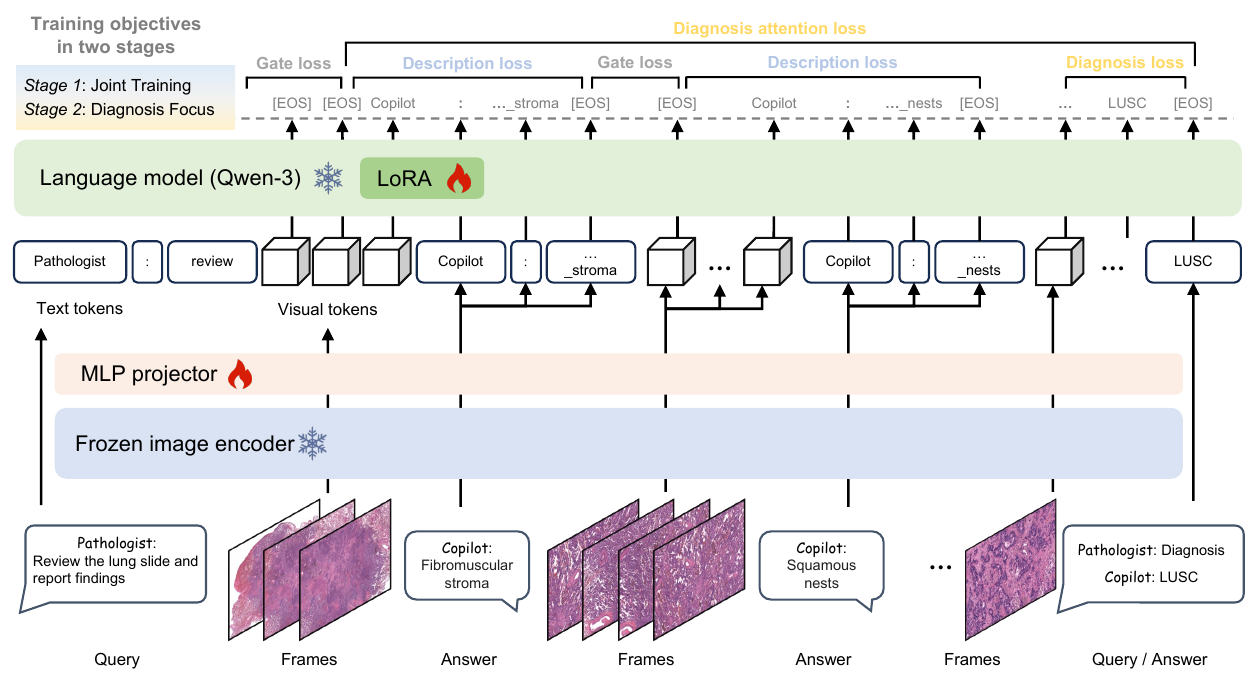}
    \caption{\textbf{PathStream overview.}
    PathStream jointly learns response timing, morphological description,
    and final diagnosis from a causal sequence of frames and responses,
    with diagnostic attention supervision guiding evidence integration.}
    \label{fig:method_overview}
\end{figure*}

\begin{figure}[t]
    \centering
    \includegraphics[width=\linewidth]{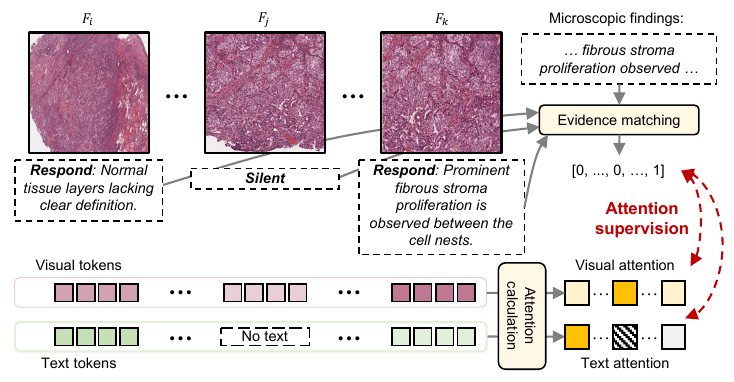}
\caption{\textbf{Diagnostic evidence attention alignment.}
    Matching descriptions to microscopic findings guides attention
    to supporting evidence during diagnosis generation.}
    \label{fig:diagnostic_attention}
\end{figure}

\subsection{Problem Formulation}
\label{sec:problem_formulation}

We define \emph{pathologist-led streaming diagnosis} as sequential
assistance during WSI review, with navigation controlled by the
pathologist. Let $X=(x_1,\ldots,x_T)$ denote the ordered frames
reconstructed from the pathologist's viewing trajectory, each
representing the field of view at that time.

At time $t$, the model uses only the frames observed so far
$x_{1:t}$, the initial assistance request $u$, and its own preceding
outputs $o_{1:t-1}$ to remain silent or generate a textual response:
\begin{equation}
    o_t=f_\theta(x_{1:t},u,o_{1:t-1})
    \in\{\text{silence}\}\cup\mathcal Y,
    \label{eq:streaming_output}
\end{equation}
where $f_\theta$ denotes the model and $\mathcal Y$ the space of
textual responses. At session end, the model produces a final
diagnosis $z$ from the observed frames and its generated
morphological descriptions.

This setting requires
\textbf{(1) response timing}: responding to new morphological
information while remaining silent during continued inspection
of similar views;
\textbf{(2) morphological interpretation}: accurately describing
morphology across tissue regions and magnification levels;
and \textbf{(3) diagnostic synthesis}: integrating relevant visual
and textual evidence across the full session.

\subsection{Trajectory Acquisition and Supervision}
\label{sec:trajectory_supervision}

\subsubsection{Pathologist-Guided Trajectory Acquisition}
\label{sec:trajectory_acquisition}

We record pathologists' WSI examinations as viewing trajectories using a customized viewer. Pathologists navigate through panning, region selection and zooming, while the viewer records timestamped viewport samples
at 10\,Hz. At the end of each review, the pathologist provides
microscopic findings and a diagnostic conclusion. All diagnoses are independently verified by a sixth, supervising pathologist.

Using this workflow, five pathologists review 1,791 WSIs spanning
10 organ categories from TCGA,
and we collect trajectories from 83 additional WSIs in a private
three-center cohort covering six of these organ categories. We complement these recordings with 90 WSIs of a new organ, skin,
from the publicly released PEAN data~\cite{nanDeepLearningQuantifies2025},
which provide viewing trajectories on a 50\,Hz timebase and diagnostic
conclusions. 
Each WSI contains one viewing trajectory. For TCGA, the private
cohort, and PEAN, we uniformly sample each trajectory at 1\,Hz
and reconstruct a video using the recorded WSI coordinates
and temporal information.
\cref{fig:benchmark_overview}(a) shows the cohorts' organ distribution.

\subsubsection{Response Annotation and Frame Grouping}
\label{sec:view_grouping}

An ideal copilot should promptly summarize key morphological
features as the pathologist shifts to a new view, guiding efficient
inspection, and remain silent while similar morphology persists.
Pathologists therefore review the reconstructed trajectory videos
and individually confirm response frames that introduce new
morphological information. They mark these frames as response
points and annotate their morphology; all other frames are
labeled as silence. 

These response points partition each trajectory into contiguous
frame groups conveying similar pathological visual information.
Each group begins with a response frame and ends immediately
before the next response frame, with the final group extending
to the end of the sequence. We represent each trajectory as:
\begin{equation}
    \mathcal T = ((G_1,d_1),\ldots,(G_K,d_K)),
    \label{eq:trajectory_representation}
\end{equation}
where $G_k=(x_{b_k},\ldots,x_{e_k})$ is the $k$-th frame group
with first and last frame indices $b_k$ and $e_k$,
$d_k$ is the morphological description of its response frame
$x_{b_k}$, and $K$ is the number of groups.
Frame-group statistics are detailed in
supplementary \cref{sec:supp_view_groups}.

\subsection{PathStreamBench}
\label{sec:benchmark}
We split the 1,791 TCGA trajectories into 988 for training,
400 for validation, and 403 for testing. We construct \textbf{PathStreamBench}
from the test split to evaluate pathologist-led streaming diagnosis
through three progressively demanding tasks: response timing,
morphological interpretation, and diagnostic synthesis
(\cref{fig:benchmark_overview}(b)).

\noindent\textbf{Task~1: Response Timing.}
This task assesses response timing across different
response--silence patterns (see supplementary \cref{sec:supp_response_patterns}). Models receive frames sequentially
and must respond to local visual changes while remaining silent
during continued inspection of visually similar frames.

\noindent\textbf{Task~2: Streaming Morphological Description.}
Building on Task~1, this task requires models to respond at
appropriate moments and accurately describe morphology across
magnification levels. Generated descriptions are compared with
pathologist annotations.

\noindent\textbf{Task~3: Session-Level Diagnosis.}
This task extends Task~2 to complete viewing sessions. Models
must provide timely, accurate morphological descriptions without
unnecessary repetition and integrate the accumulated findings
into a final diagnosis at session end.
We additionally include 500 public educational pathology videos
from YouTube~\cite{vuongVideoPathLLaVAPathologyDiagnostic2025} to assess generalizations to educational slide reviews.

\noindent\textbf{Task~4: Pathology Video Understanding.}
This auxiliary task complements streaming diagnosis by assessing
recognition and temporal reasoning. Models answer multiple-choice
questions about visual content, temporal change, and temporal
order in trajectory clips.

\cref{fig:benchmark_overview}(c) summarizes the sample distribution:
500 cases each for Tasks~1 and~2, 403 WSI trajectories and
500 videos for Task~3, and 420 MCQs for Task~4.
\section{PathStream}
\label{sec:pathstream}
\subsection{Model Overview}
\label{sec:model_overview}
\Cref{fig:method_overview} illustrates PathStream, a copilot
for pathologist-led streaming diagnosis. During review, the model outputs description as the view
changes and remains silent during continued inspection of similar
frames. The observed frames and generated descriptions form the
visual and textual history used for final diagnosis. We jointly supervise streaming responses and final diagnoses,
with a diagnosis-attention loss guiding the use of historical evidence.

Built on Qwen3-VL-4B-Instruct~\cite{baiQwen3VLTechnicalReport2025} and VideoLLM-online \cite{chenVideoLLMonlineOnlineVideo2024a}
PathStream is trained on complete chronological video sequences.
Each sequence interleaves frames with annotated responses and ends
with the pathologist-verified diagnosis.
We apply LoRA~\cite{huLoRALowRankAdaptation2022} ($r=128$)
to the model's attention and feed-forward layers
and train the adapters together with the vision--language merger
MLPs while freezing all other parameters. Implementation details
in \cref{sec:supp_implementation}.

% \subsection{Model Overview}
% \label{sec:model_overview}
% \Cref{fig:method_overview} illustrates the PathStream architecture
% and training objectives. Built on
% Qwen3-VL-4B-Instruct~\cite{baiQwen3VLTechnicalReport2025},
% PathStream learns from complete streaming sequences constructed
% from annotated trajectories. Each sequence begins with a user request for assistance,
% followed by frames in chronological order. The response-control
% loss uses frame-level response labels to teach the model to respond or
% remain silent, while the morphological description loss supervises
% responses with annotated descriptions. At session end, the
% diagnosis generation loss supervises diagnostic synthesis using
% the pathologist-provided diagnosis. The diagnostic attention loss
% guides attention to supporting frames and descriptions identified
% from microscopic findings.

% We use two-stage supervised fine-tuning: Stage~1 jointly learns
% streaming interaction and diagnosis, while Stage~2 emphasizes
% diagnostic synthesis. We apply LoRA~\cite{huLoRALowRankAdaptation2022}
% ($r=128$) to the language model's attention and feed-forward layers,
% jointly training the adapters and vision--language merger MLPs
% while freezing all other parameters. Implementation details
% appear in \cref{sec:supp_implementation}.

\subsection{Streaming Generation with Response Control}
\label{sec:streaming_generation}
\label{sec:morphological_description}

PathStream learns to describe morphology as the view changes
and remain silent during continued inspection of similar frames.
Response labels encode pathologist-confirmed morphological
changes: each frame group's first frame is marked for response,
while subsequent similar frames are marked for silence.
After each frame, the gate loss $\mathcal L_g$ supervises
end-of-sequence (EOS) token
prediction~\cite{chenVideoLLMonlineOnlineVideo2024a}:
EOS indicates silence, whereas a non-EOS token initiates a response.
At annotated response points, the description loss $\mathcal L_d$
supervises description generation using the pathologist's
annotation of the current frame.

% \subsection{Streaming Generation with Response Control}
% \label{sec:streaming_generation}

% We follow the LIVE formulation of
% VideoLLM-online~\cite{chenVideoLLMonlineOnlineVideo2024a} to train the
% model to respond to visual changes without repeatedly describing
% similar consecutive frames. The gate loss $\mathcal L_g$ supervises
% end-of-sequence (EOS) token prediction after each frame: EOS indicates
% silence, while a non-EOS prediction triggers a response.

% \subsection{Morphological Description}
% \label{sec:morphological_description}

% Once a response is triggered, we supervise description generation with
% the pathologist's annotation of the current frame. The description loss
% $\mathcal L_d$ encourages accurate descriptions of the observed
% morphology.

\begin{table*}[t]
    \centering
    \caption{\textbf{Trajectory and video benchmark results.}
    Tasks~1, 2, and~4 use trajectory inputs.
    All scores are multiplied by 100; higher is better.
    \textbf{Bold} and \underline{underlined} scores indicate
    the best and second-best reported results in each column.}
    \label{tab:benchmark_results}
    \setlength{\tabcolsep}{2pt}
    \begin{tabular*}{\linewidth}{@{\extracolsep{\fill}}l*{8}{c}@{}}
        \toprule
        & Task 1 & \multicolumn{2}{c}{Task 2}
        & \multicolumn{4}{c}{Task 3} & Task 4 \\
        & Gate & \multicolumn{2}{c}{Description}
        & \multicolumn{4}{c}{Diagnosis} & QA \\
        \cmidrule(lr){2-2}\cmidrule(lr){3-4}
        \cmidrule(lr){5-8}\cmidrule(lr){9-9}
        Model & Macro-F1 & ROUGE-L & CONCH
        & \shortstack{Video-\\Precision} & \shortstack{Video-\\Relevance}
        & \shortstack{Trajectory-\\Precision} & \shortstack{Trajectory-\\Relevance} & Accuracy \\
        \midrule
        \rowcolor{modelgroupblue}
        \multicolumn{9}{@{}l@{}}{\textit{General VLMs}} \\
        InternVL3.5-8B & 64.86 & 13.70 & 27.50 & 10.73 & 12.67 & 16.60 & 30.74 & 48.33 \\
        Qwen3-VL-4B-Instruct & 64.71 & 11.22 & 22.73 & 8.26 & 5.29 & 14.67 & 9.29 & 49.29 \\
        Qwen3-VL-8B-Instruct & 57.14 & 0.45 & 0.69 & 9.03 & 8.03 & 16.84 & 18.73 & 45.00 \\
        GPT-6-luna & 78.76 & 18.38 & 36.16 & 15.25 & 10.14 & 17.76 & 15.65 & 50.71 \\
        Qwen3.6-35B-A3B & 43.75 & 6.69 & 14.92 & 17.46 & 18.22 & 20.88 & 25.28 & \underline{51.43} \\
        Qwen3.8-Flash & 45.21 & 15.61 & 32.97 & 18.83 & 22.10 & 19.54 & 30.98 & \textbf{52.14} \\
        \midrule
        \rowcolor{modelgroupblue}
        \multicolumn{9}{@{}l@{}}{\textit{Streaming VLMs}} \\
        \shortstack[l]{VideoLLM-online} & 28.91 & 1.30 & 6.31 & 1.32 & 2.38 & 2.32 & 3.58 & 11.19 \\
        StreamingVLM & 49.78 & 2.18 & 5.25 & 0.92 & 1.09 & 1.44 & 2.63 & 35.48 \\
        LiveCC-7B-Instruct & 39.71 & 4.30 & 12.05 & 5.23 & 8.14 & 8.42 & 12.58 & 42.62 \\
        AURA & 32.28 & 1.55 & 2.01 & 5.56 & 5.62 & 20.74 & 23.62 & 37.62 \\
        MOSS-VL-Realtime & 48.96 & 5.31 & 9.89 & 6.22 & 4.45 & 9.80 & 8.92 & 33.10 \\
        MMDuet2 & 51.23 & 4.85 & 8.64 & 0.57 & 0.80 & 3.13 & 2.88 & 40.48 \\
        \midrule
        \rowcolor{modelgroupblue}
        \multicolumn{9}{@{}l@{}}{\textit{Pathology / Medical VLMs}} \\
        Patho-R1-7B & 39.90 & 11.15 & 22.34 & 16.32 & 11.58 & 20.16 & 14.17 & 42.14 \\
        Lingshu-32B & 51.49 & 5.09 & 18.64 & 10.31 & 11.17 & 15.23 & 27.52 & 50.48 \\
        \midrule
        \rowcolor{modelgroupblue}
        \multicolumn{9}{@{}l@{}}{\textit{Domain SFT baselines}} \\
        LiveCC-7B + SFT & 88.60 & 26.51 & 47.33 & 36.62 & \underline{37.78} & 56.45 & 60.34 & 50.00 \\
        MMDuet2 + SFT & 89.70 & 25.32 & 45.58 & \underline{36.68} & 36.97 & 54.66 & 58.90 & 48.33 \\
        Qwen3-VL-4B + SFT & \underline{93.97} & \underline{27.82} & \underline{48.35} & 32.46 & 33.56 & \underline{60.50} & \underline{62.76} & 50.00 \\
        \midrule
        \textbf{PathStream} & \textbf{95.10} & \textbf{31.27} & \textbf{51.55} & \textbf{48.55} & \textbf{50.36} & \textbf{75.22} & \textbf{77.87} & \textbf{52.14} \\
        \bottomrule
    \end{tabular*}
\end{table*}

\subsection{Diagnostic Evidence Alignment}
\label{sec:diagnostic_evidence_alignment}

Streaming responses describe frame-level morphology, whereas
session-level diagnosis requires integrating relevant findings
across the accumulated frames and descriptions. At session end,
the model generates a diagnosis upon request. The diagnosis loss
$\mathcal L_z$ supervises this output using the pathologist-verified
diagnosis. To guide diagnosis generation, we construct smoothed attention
targets~\cite{huAttendEvidenceEvidenceAnchored2026}
over supporting historical frames and descriptions.

As illustrated in \cref{fig:diagnostic_attention}, a language model
matches annotated frame descriptions to the pathologist-provided
microscopic findings. Each matched frame group contributes its
response frame and description as a supporting evidence pair,
linking session-level findings to specific observations. These pairs define separate targets $P_z^m$ over historical
visual ($m=v$) and description ($m=t$) tokens.

Let $A_u^m$ denote the model's attention distribution for diagnosis
token $u$. We align this distribution with the evidence target
by minimizing
\begin{equation}
    \mathcal L_{\mathrm{za}}
    = \mathbb E_{m,u}
    \left[D_{\mathrm{KL}}\!\left(A_u^m\,\|\,P_z^m\right)\right].
    \label{eq:diag_attention_loss}
\end{equation}
Evidence matching, target construction, and attention computation
are detailed in supplementary \cref{sec:supp_diagnostic_attention}.

\subsection{Training Objective and Optimization}
\label{sec:training}

We use token-level cross-entropy for $\mathcal L_g$,
$\mathcal L_d$, and $\mathcal L_z$:
\begin{equation}
    \mathcal L_r =
    -\frac{1}{|\mathcal I_r|}
    \sum_{u\in\mathcal I_r}
    \log p_\theta(y_u\mid h_u),
    \label{eq:generation_loss}
\end{equation}
where $r\in\{g,d,z\}$, $\mathcal I_r$ contains the supervised token
positions, and $y_u$ and $h_u$ denote the target token and its
preceding visual and textual context. We combine these losses with
diagnostic attention supervision to form the overall objective at
training stage $s\in\{1,2\}$:
\begin{equation}
    \mathcal L^{(s)} =
    w_g^{(s)}\mathcal L_g
    + w_d^{(s)}\mathcal L_d
    + w_z^{(s)}\mathcal L_z
    + \lambda_z\mathcal L_{\mathrm{za}}.
    \label{eq:training_total_loss}
\end{equation}

We train on complete chronological sessions in two stages.
Stage~1 jointly trains the model for response control,
morphological description, and diagnostic synthesis.
Stage~2 continues training with greater emphasis on diagnosis
while retaining the response-control and description objectives.
Token-level supervision, loss weights, and optimization settings
are detailed in supplementary
\cref{sec:supp_token_supervision,sec:supp_optimization}.

\section{Experiments}
\label{sec:experiments}
\subsection{Experimental Setup}
\label{sec:experimental_setup}

\noindent\textbf{Comparison models.}
We train both PathStream stages on 988 training trajectories
and select loss weights and LoRA rank using 400 validation trajectories.
We benchmark PathStream against 14 general, streaming,
and pathology/medical VLMs on the 403 held-out test trajectories.
The six \emph{general VLMs} are
InternVL3.5-8B~\cite{wangInternVL35AdvancingOpenSource2025},
Qwen3-VL-4B-Instruct and
Qwen3-VL-8B-Instruct~\cite{baiQwen3VLTechnicalReport2025},
GPT-6-luna~\cite{IntroducingGPT6Sol2026},
Qwen3.6-35B-A3B~\cite{Qwen3.6-35B-A3B}, and
Qwen3.8-Flash~\cite{Qwen38flashAlibabaCloud}.
The six \emph{streaming VLMs} are
VideoLLM-online-8B-v1plus~\cite{chenVideoLLMonlineOnlineVideo2024a},
StreamingVLM~\cite{xuStreamingVLMRealTimeUnderstanding2026a},
LiveCC-7B-Instruct~\cite{chenLiveCCLearningVideo2025},
AURA~\cite{luAURAAlwaysOnUnderstanding2026},
MOSS-VL-Realtime~\cite{wangMOSSVLTechnicalReport2026}, and
MMDuet2~\cite{wangMMDuet2EnhancingProactive2026}.
\emph{Pathology/medical VLMs} include
Patho-R1-7B~\cite{zhangPathoR1MultimodalReinforcement2026b}
and Lingshu-32B~\cite{xuLingshuGeneralistFoundation2026a}. We also fine-tune LiveCC-7B, MMDuet2, and Qwen3-VL-4B
(whose SFT variant is also reported as ``w/o diagnosis attention''
in the Stage~1 ablation) on our training data as domain SFT baselines
(supplementary \cref{sec:supp_full_results}).

\noindent\textbf{Evaluation protocol.}
General and pathology/medical VLMs receive each frame with
earlier frames and their own generated descriptions, and output
either \texttt{silent} or a morphological description.
Streaming VLMs use their native response mechanisms and
streaming state. Task~3 ends with a diagnosis request.
For Task~4, models receive a video clip and answer a
multiple-choice question. Details are provided in supplementary
\cref{sec:supp_evaluation_protocol,sec:supp_prompts}.

\noindent\textbf{Evaluation metrics.}
For Task~1, gate Macro-F1 averages the F1 scores for response
and silence. Task~2 uses ROUGE-L~\cite{linROUGEPackageAutomatic2004}
and CONCH text similarity~\cite{luVisuallanguageFoundationModel2024b}.
For Task~3, we use GPT-6-luna to evaluate WSI Precision
and WSI Relevance~\cite{liangWSILLaVAMultimodalLarge2025}.
WSI Precision measures how much of the reference diagnosis is
correctly captured, while WSI Relevance measures how well the
generated diagnosis agrees with the reference.
Task~4 uses accuracy. Diagnosis scoring is detailed in
supplementary \cref{sec:supp_diagnosis_metrics}. Higher values indicate better performance for all metrics. Pathologists additionally reviewed a subset of generated descriptions
and diagnoses, and their qualitative assessments were consistent
with the CONCH, WSI Precision, and WSI Relevance scores.

\begin{table*}[t]
    \centering
    \caption{\textbf{Ablation results and training-stage comparison.}
    PathStream (Stage 1) is the reference for the four objective ablations.
    PathStream (Stage 2) reports the complete two-stage model.
    Tasks~1, 2, and~4 use trajectory inputs.
    All scores are multiplied by 100; higher is better.
    \textbf{Bold} and \underline{underlined} scores indicate
    the best and second-best reported results in each column.}
    \label{tab:ablation_results}
    \setlength{\tabcolsep}{2pt}
    \begin{tabular*}{\linewidth}{@{\extracolsep{\fill}}l*{8}{c}@{}}
        \toprule
        & Task 1 & \multicolumn{2}{c}{Task 2}
        & \multicolumn{4}{c}{Task 3} & Task 4 \\
        & Gate & \multicolumn{2}{c}{Description}
        & \multicolumn{4}{c}{Diagnosis} & QA \\
        \cmidrule(lr){2-2}\cmidrule(lr){3-4}
        \cmidrule(lr){5-8}\cmidrule(lr){9-9}
        Model & Macro-F1 & ROUGE-L & CONCH
        & \shortstack{Video-\\Precision} & \shortstack{Video-\\Relevance}
        & \shortstack{Trajectory-\\Precision} & \shortstack{Trajectory-\\Relevance} & Accuracy \\
        \midrule
        \textbf{PathStream (Stage 1)} & \underline{94.61} & 28.20 & 48.75 & \underline{38.07} & \underline{39.08} & \underline{66.03} & \underline{68.22} & \textbf{54.05} \\
        \textbf{PathStream (Stage 2)} & \textbf{95.10} & \textbf{31.27} & \textbf{51.55} & \textbf{48.55} & \textbf{50.36} & \textbf{75.22} & \textbf{77.87} & \underline{52.14} \\
        \midrule
        w/o diagnosis attention & 93.97 & 27.82 & 48.35 & 32.46 & 33.56 & 60.50 & 62.76 & 50.00 \\
        w/o gate CE & 37.24 & \underline{30.64} & \underline{51.25} & 32.41 & 33.75 & 62.41 & 63.13 & 50.00 \\
        w/o description CE & 92.43 & 3.34 & 25.98 & 31.80 & 32.17 & 37.90 & 34.47 & 50.00 \\
        w/o diagnosis CE & 94.05 & 27.37 & 48.11 & 7.92 & 8.38 & 1.43 & 4.64 & \underline{52.14} \\
        \bottomrule
    \end{tabular*}
\end{table*}

% Start the analysis figures on the ablation-table page.
\suppressfloats[t]
\begin{figure}[t]
    \centering
    \includegraphics[width=\linewidth,trim=0 22bp 0 38bp,clip]{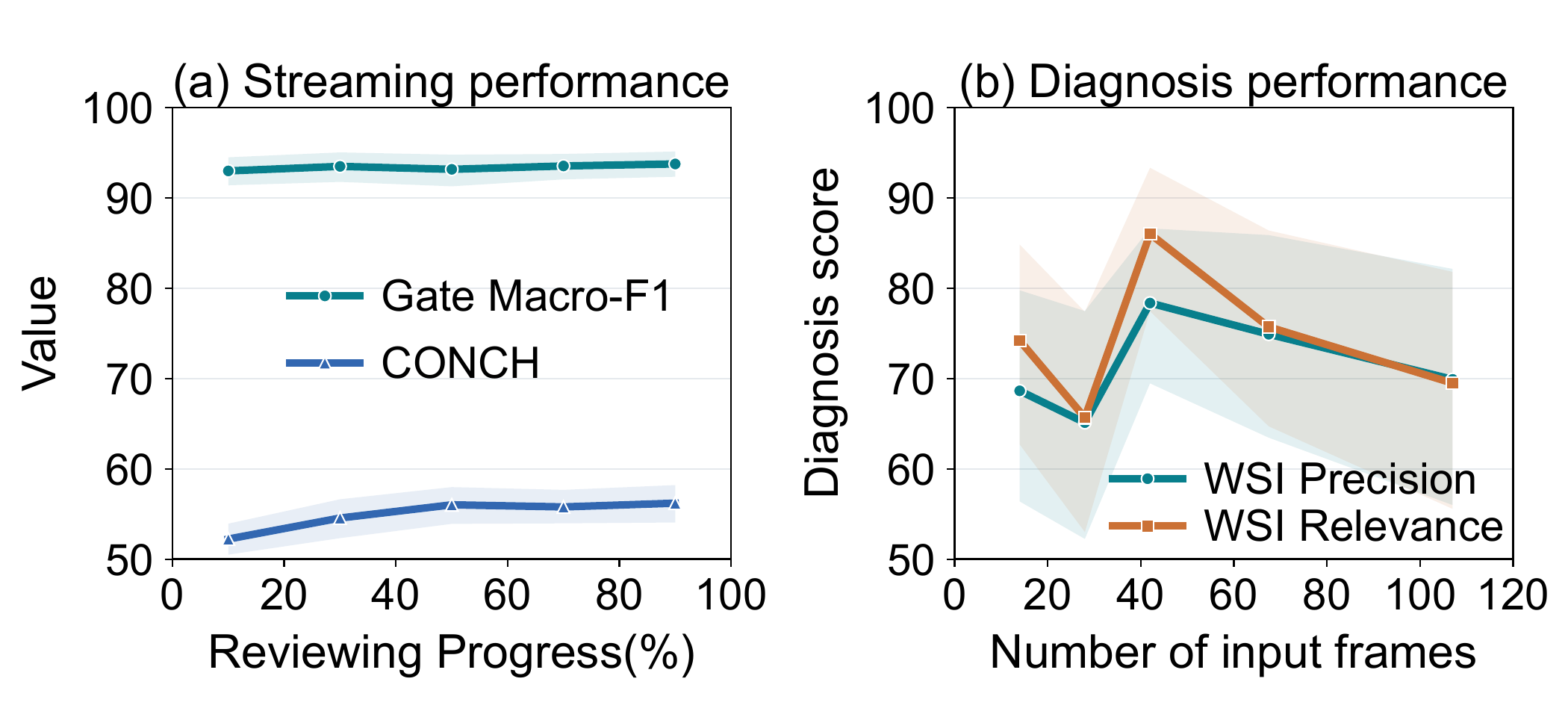}
   \caption{\textbf{Streaming analysis.}
(a) Response timing and description quality during review.
(b) Diagnosis quality by session length.}
    \label{fig:streaming_analysis}
\end{figure}

\subsection{Comparative Results}
\label{sec:main_results}
\cref{tab:benchmark_results} shows that PathStream better supports
pathologist-led streaming diagnosis than the 14 models evaluated
with their original checkpoints. PathStream provides timely updates
as the pathologist's view changes, achieving 95.10 gate Macro-F1 in
Task~1, 16.34 points above the strongest baseline. It also interprets
morphology and integrates accumulated evidence into a diagnosis,
reaching 51.55 CONCH in Task~2 and 77.87 trajectory WSI Relevance
in Task~3, surpassing the strongest baselines by 15.39 and 46.89
points, respectively. Diagnostic synthesis further generalizes to
educational videos, with 50.36 WSI Relevance, a 28.26-point
improvement over the strongest baseline. For visual and temporal
understanding, PathStream ties with Qwen3.8-Flash for the highest
Task~4 accuracy at 52.14\%.
PathStream therefore provides more effective assistance during
review.

These comparisons highlight distinct gaps in existing models:
general VLMs struggle with diagnostic synthesis, streaming VLMs
have limited morphological understanding, and pathology/medical
VLMs lack reliable response timing during sequential review.
Fine-tuning LiveCC-7B, MMDuet2, and Qwen3-VL-4B on our data
substantially improves all three capabilities, narrowing the gaps
in response timing and morphological description. PathStream
nevertheless retains an advantage, particularly in diagnostic
synthesis: it exceeds the strongest SFT baselines by 1.13 points
in gate Macro-F1 and 3.20 points in description CONCH, with larger
gains of 15.11 and 12.58 points in trajectory and video
WSI Relevance, respectively.

\begin{figure}[t]
    \centering
    \includegraphics[width=0.9\linewidth,trim=0 55bp 0 30bp,clip]{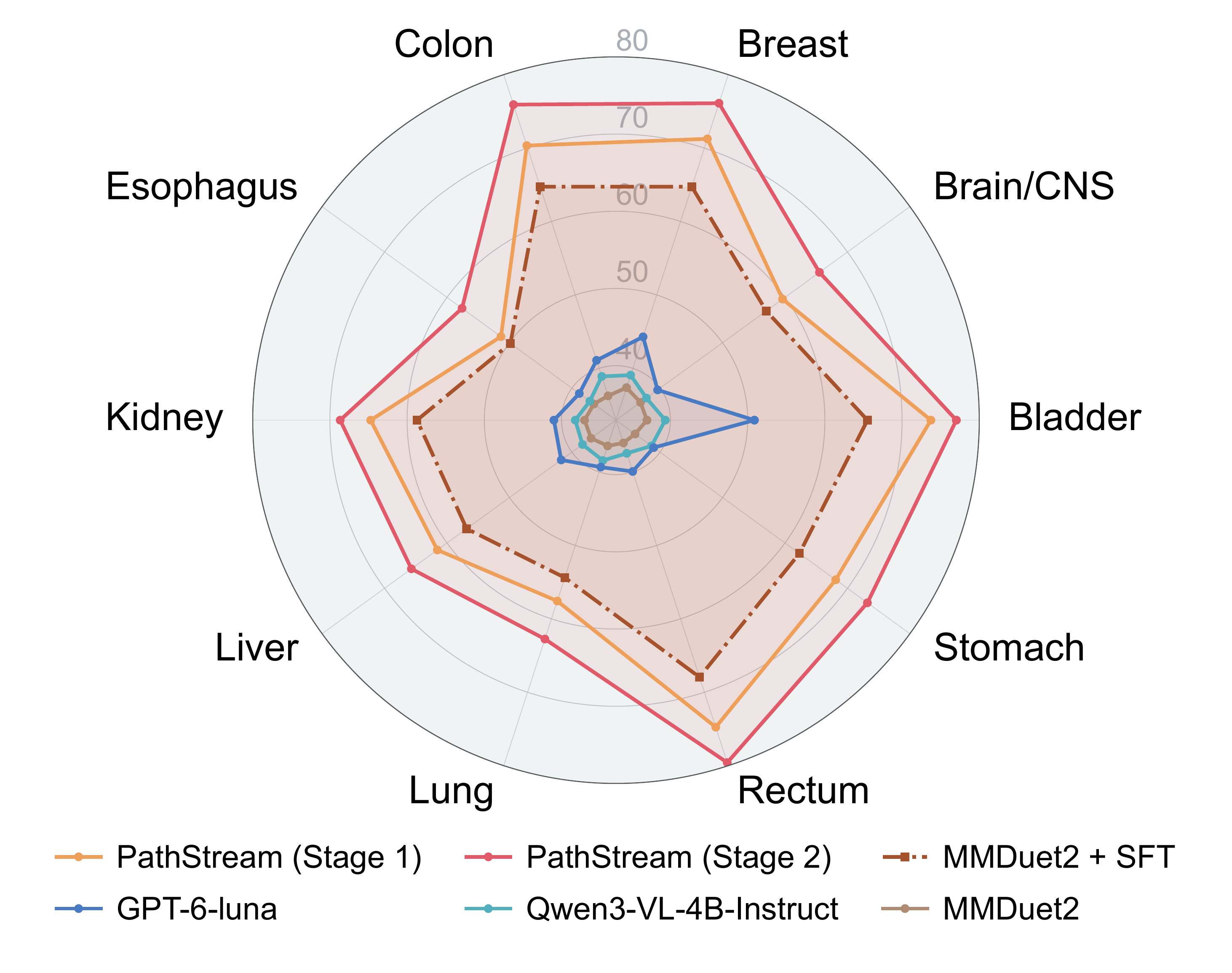}
    \caption{\textbf{Organ-wise comparison.}
    Four-task performance across ten TCGA organs.}
    \label{fig:organ_radar}
\end{figure}

\cref{fig:streaming_analysis} shows that response timing and
description quality remain stable throughout review, while
PathStream maintains WSI Precision and WSI Relevance above 65
across the evaluated session lengths. Its advantage also holds across all ten TCGA
organs (\cref{fig:organ_radar}). Domain SFT improves MMDuet2
across these organs, while PathStream achieves higher overall
scores at Stage~1 and further improves at Stage~2.
Details of organ-wise score aggregation and full results
are provided in the supplementary material
(\cref{tab:supp_benchmark_results,fig:supp_organ_radar}).

\begin{figure*}[t]
    \centering
    \includegraphics[width=\linewidth,trim=0 14bp 0 42bp,clip]{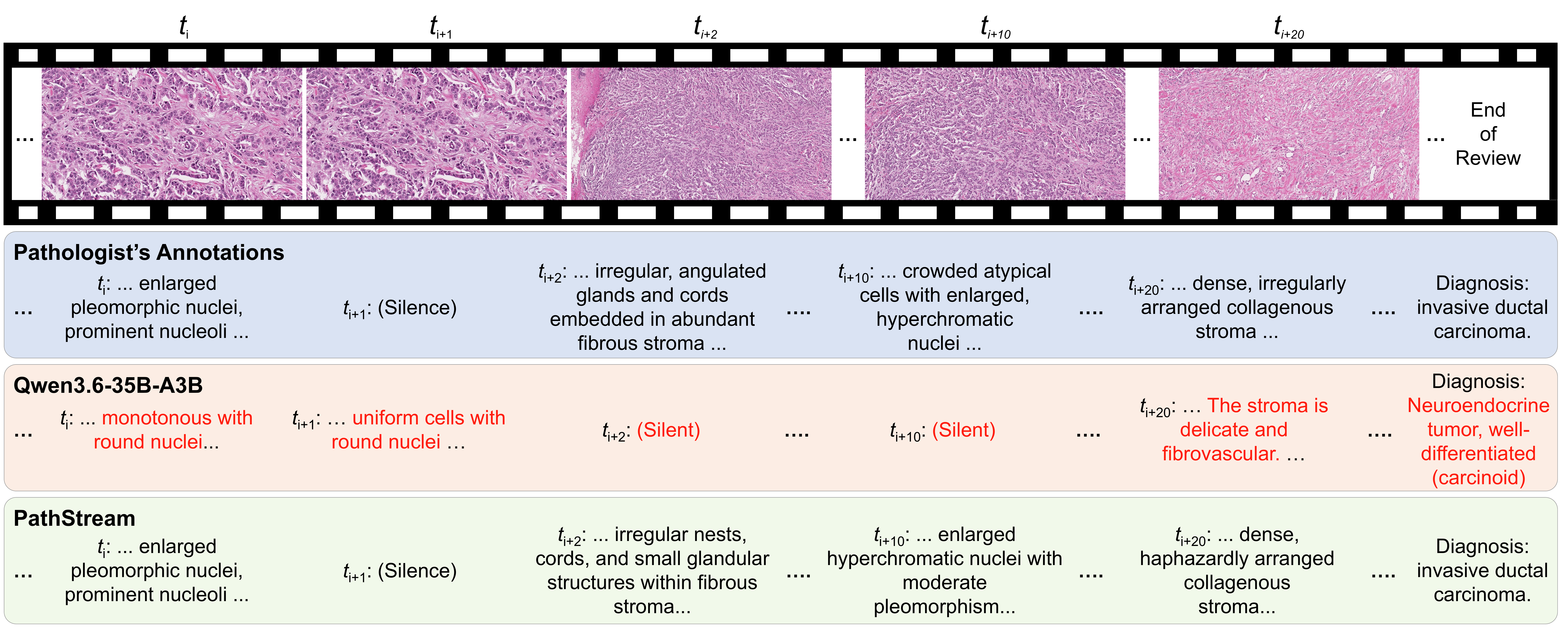}
   \caption{\textbf{Streaming responses and final diagnoses.}
Qwen3.6-35B-A3B and PathStream outputs alongside pathologist
annotations for the same WSI. {\color{red}{Red}} text marks errors in descriptions,
response timing, or diagnosis. Ellipses mark omitted frames or text.}
    \label{fig:qualitative}
\end{figure*}

\subsection{Ablation Studies}
\label{sec:ablations}

We individually ablate the four losses in \cref{eq:training_total_loss},
using PathStream Stage~1 as the reference
(\cref{tab:ablation_results}). See \cref{tab:supp_ablation_results} for full results and
\cref{sec:supp_hyperparameter_tuning} for loss-weight and
LoRA-rank ablations.

\noindent\textbf{Diagnostic attention primarily benefits final diagnosis.}
Removing diagnosis-attention supervision lowers trajectory WSI Precision
and Relevance by 5.53 and 5.46 points, while gate and description scores
each change by less than one point. Diagnosis also declines on educational
pathology videos. This selective effect supports its role in focusing diagnosis
on relevant historical evidence.

\noindent\textbf{Joint supervision supports pathologist-led streaming diagnosis.}
Removing gate supervision lowers gate Macro-F1 from 94.61 to 37.24,
while description quality is preserved. Removing description
supervision reduces Task~2 ROUGE-L from 28.20 to 3.34 and CONCH
from 48.75 to 25.98. Trajectory WSI Relevance also falls from
68.22 to 34.47 even with diagnosis supervision retained.
Removing diagnosis supervision lowers trajectory WSI Relevance
to 4.64, while response timing and description scores remain
nearly unchanged. These results reveal complementary roles for
the three objectives. Accurate morphological descriptions alone
ensure neither appropriate response timing nor effective
session-level diagnosis. Gate supervision teaches the model when
to contribute during review, while morphological supervision
supports both local interpretation and the observations used
for diagnostic synthesis. Diagnosis supervision further teaches
the model to integrate these observations into a conclusion.
Joint learning thus connects timely morphological assistance
during the examination with diagnosis from accumulated evidence.

\noindent\textbf{Stage~2 strengthens diagnosis while maintaining streaming performance.}
Stage~2 improves trajectory WSI Precision and Relevance by 9.19 and
9.65 points, respectively. These gains show that increasing the weight
of sparse diagnosis supervision improves diagnostic synthesis while
maintaining response timing and description quality.

\begin{figure}[t]
    \centering
    \includegraphics[width=\linewidth]{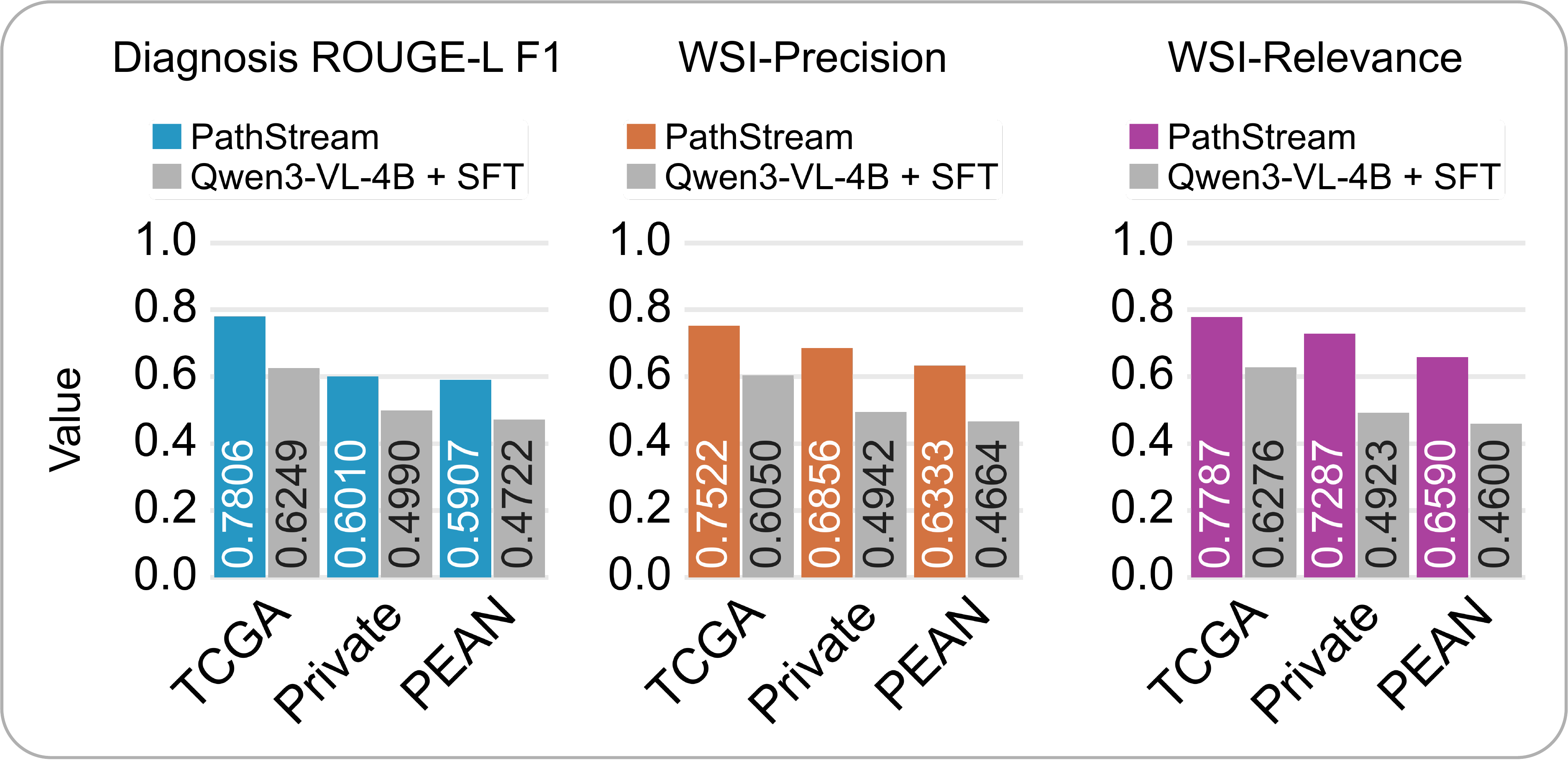}
    \caption{\textbf{External diagnostic generalization.}
PathStream and domain-fine-tuned Qwen3-VL-4B on the TCGA test set,
the private three-center cohort, and the PEAN skin cohort.
Left to right: diagnosis ROUGE-L F1, WSI Precision, and WSI Relevance.}
    \label{fig:external_validation}
\end{figure}

\subsection{External Validation and Generalization}
We evaluate TCGA-trained PathStream on the private three-center
and PEAN skin cohorts without further training.
\cref{fig:external_validation} shows that PathStream outperforms
domain-fine-tuned Qwen3-VL-4B across all three metrics.
It achieves 72.87 WSI Relevance on the private cohort,
23.64 points above the baseline. On PEAN, it achieves
63.33 WSI Precision and 65.90 WSI Relevance, with gains of
16.69 and 19.90 points, respectively.
Both models perform worse outside TCGA, but PathStream shows
smaller declines in WSI Precision and Relevance on both cohorts. Its WSI Relevance advantage on both external cohorts also
exceeds the 15.11-point gain observed on TCGA.
It therefore better preserves diagnostic quality across clinical
centers and extends diagnostic synthesis to an organ absent
from our training data.

\subsection{Qualitative Analysis}
\label{sec:qualitative_analysis}
\cref{fig:qualitative} compares Qwen3.6-35B-A3B and PathStream against
pathologist annotations from the same WSI viewing sequence.
PathStream follows the annotated response--silence pattern,
describes the changing morphology, and reaches the reference
diagnosis of invasive ductal carcinoma. In contrast, Qwen3.6-35B-A3B misinterprets morphology,
responds unnecessarily, misses relevant updates, and
incorrectly diagnoses a well-differentiated neuroendocrine tumor.
This example shows how PathStream supports morphological
interpretation and diagnosis as visual evidence accumulates
across successive views during pathologist-led review.
Full responses appear in supplementary \cref{fig:qualitative_full}.
\section{Conclusion}
\label{sec:conclusion}

Here, we formalize pathologist-led streaming diagnosis, in which
a copilot follows the pathologist's examination and integrates evidence
as it accumulates. We collect annotated WSI viewing trajectories and establish PathStreamBench to evaluate response timing, morphological interpretation, and session-level diagnosis. We further design PathStream, which delivers timely morphological descriptions, avoids redundant responses, and integrates the evidence into a final diagnosis. Experiments show that PathStream achieves superior performance and generalizes to new clinical centers and an organ unseen during training. PathStreamBench and PathStream lay the foundation for pathology copilots that support diagnosis as evidence unfolds.

% User-maintained Zotero export. Cite the keys present in main.bib.
{
    \small
    \bibliographystyle{ieeenat_fullname}
    \bibliography{main}
}

\clearpage
\appendix
% Let supplementary text flow continuously around standard full-width floats.
\twocolumn[{
\begin{center}
    \Large\textbf{\thetitle}\\
    \medskip Supplementary Material
\end{center}
}]
\raggedbottom
\setcounter{dbltopnumber}{1}
\renewcommand{\dblfloatpagefraction}{0.9}
\makeatletter
\setlength{\@dblfptop}{0pt}
\makeatother
\microtypesetup{expansion=true}
\brokenpenalty=10000

\setcounter{table}{0}
\renewcommand{\thetable}{S\arabic{table}}
\renewcommand{\theHtable}{supp.\arabic{table}}

\setcounter{figure}{0}
\renewcommand{\thefigure}{S\arabic{figure}}
\renewcommand{\theHfigure}{supp.\arabic{figure}}

\noindent\textbf{Table of Contents}
\begin{itemize}
    \interlinepenalty=10000
    \item \hyperref[sec:supp_cohort]{\S\ref*{sec:supp_cohort}: Cohort Composition}
    \item \hyperref[sec:supp_view_groups]{\S\ref*{sec:supp_view_groups}: Trajectory Processing and Frame Grouping}
    \item \hyperref[sec:supp_implementation]{\S\ref*{sec:supp_implementation}: Implementation Details}
    \begin{itemize}
        \item \hyperref[sec:supp_token_supervision]{\S\ref*{sec:supp_token_supervision}: Token-Level Supervision}
        \item \hyperref[sec:supp_diagnostic_attention]{\S\ref*{sec:supp_diagnostic_attention}: Diagnostic Attention Loss}
        \item \hyperref[sec:supp_optimization]{\S\ref*{sec:supp_optimization}: Two-Stage Optimization}
    \end{itemize}
    \item \hyperref[sec:supp_benchmark_details]{\S\ref*{sec:supp_benchmark_details}: PathStreamBench Details}
    \begin{itemize}
        \item \hyperref[sec:supp_evaluation_protocol]{\S\ref*{sec:supp_evaluation_protocol}: Evaluation Protocol}
        \item \hyperref[sec:supp_prompts]{\S\ref*{sec:supp_prompts}: Prompts and Conversation Format}
        \item \hyperref[sec:supp_diagnosis_metrics]{\S\ref*{sec:supp_diagnosis_metrics}: WSI Precision and WSI Relevance}
        \item \hyperref[sec:supp_response_patterns]{\S\ref*{sec:supp_response_patterns}: Task~1 Response--Silence Patterns}
    \end{itemize}
    \item \hyperref[sec:supp_full_results]{\S\ref*{sec:supp_full_results}: Full Benchmark Results}
    \item \hyperref[sec:supp_hyperparameter_tuning]{\S\ref*{sec:supp_hyperparameter_tuning}: Hyperparameter Ablations}
    \item \hyperref[sec:supp_organ_results]{\S\ref*{sec:supp_organ_results}: Organ-Wise Benchmark Results}
    \item \hyperref[sec:supp_qualitative]{\S\ref*{sec:supp_qualitative}: Complete Qualitative Responses}
    \item \hyperref[sec:supp_clinical_case]{\S\ref*{sec:supp_clinical_case}: Multi-Pathologist Case Study}
\end{itemize}

\section{Cohort Composition}
\label{sec:supp_cohort}

\Cref{tab:data_sources} provides the WSI counts underlying the organ
distribution in \cref{fig:benchmark_overview}(a).
\begin{table*}[t]
    \centering
    \caption{WSI counts by source and organ. The private cohort and
    PEAN are reserved for external validation of PathStream.
    Dashes indicate organs not represented in a source.}
    \label{tab:data_sources}
    \begin{tabular*}{\linewidth}{@{\extracolsep{\fill}}lrrrrrr@{}}
        \toprule
        & \multicolumn{4}{c}{TCGA} & \multicolumn{2}{c}{External} \\
        \cmidrule(lr){2-5}\cmidrule(l){6-7}
        Organ & Total & Train & Validation & Test & Private & PEAN \\
        \midrule
        Bladder   & 195 & 108 & 43 & 44 & -- & -- \\
        Brain/CNS & 295 & 163 & 66 & 66 & -- & -- \\
        Breast    & 210 & 116 & 47 & 47 & 15 & -- \\
        Colon     & 138 &  76 & 31 & 31 & 15 & -- \\
        Esophagus &  40 &  22 &  9 &  9 & -- & -- \\
        Kidney    & 338 & 186 & 76 & 76 & -- & -- \\
        Liver     & 172 &  95 & 38 & 39 & 15 & -- \\
        Lung      & 278 & 153 & 62 & 63 & 15 & -- \\
        Rectum    &  56 &  31 & 13 & 12 &  8 & -- \\
        Skin      &  -- &  -- & -- & -- & -- & 90 \\
        Stomach   &  69 &  38 & 15 & 16 & 15 & -- \\
        \midrule
        Total & 1,791 & 988 & 400 & 403 & 83 & 90 \\
        \bottomrule
    \end{tabular*}
\end{table*}

\section{Trajectory Processing and Frame Grouping}
\label{sec:supp_view_groups}

Before constructing model inputs, we exclude unusually long sessions
dominated by prolonged pauses at a fixed viewport, typically caused by
interruptions in the pathologist's examination.

\begin{figure}[!htb]
    \centering
    \includegraphics[width=\linewidth]{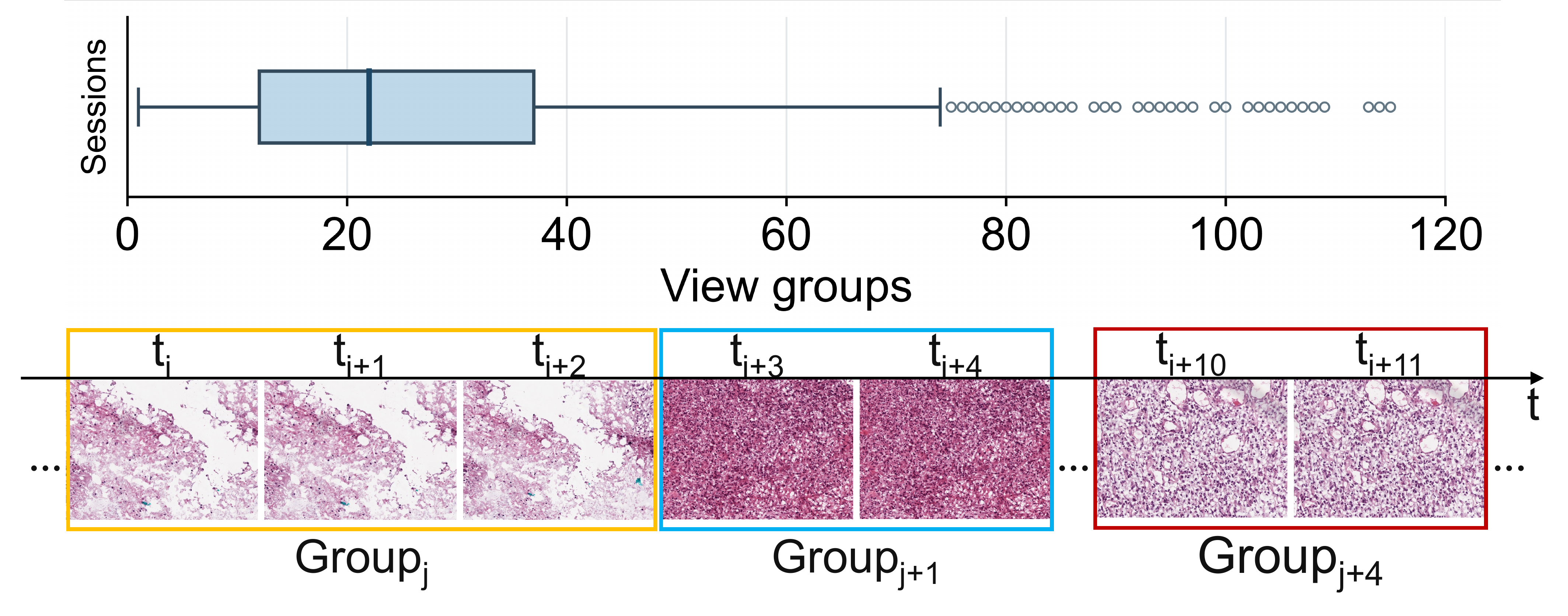}
    \caption{\textbf{Frame-group counts across sessions and group boundaries
    in an example trajectory.}}
    \label{fig:view_group_distribution}
\end{figure}

Viewing sessions contain a median of 22 frame groups
(interquartile range 12--37), as shown in the top panel of
\cref{fig:view_group_distribution}. In the example trajectory below
the box plot, frames $t_i$ through $t_{i+2}$ show similar morphology
and belong to group $G_j$. At $t_{i+3}$, new morphology appears,
and the pathologist marks this frame as the response point that starts
group $G_{j+1}$. Frame $t_{i+4}$ shows similar morphology and remains
in the same group.

\section{Implementation Details}
\label{sec:supp_implementation}

\paragraph{Visual inputs and context.}
PathStream processes frames chronologically, adding one new frame
at each streaming step. The number of input frames varies with
the clip or session length.
Images are resized with their aspect ratios preserved under a
budget of 100,352 pixels per frame. With $16\times16$ patches
and $2\times2$ spatial merging, each frame contributes at most
98 visual tokens, giving at most $98T$ visual tokens for
a sequence of $T$ frames.
The context budget is 24,576 tokens, shared by visual inputs,
instructions, and textual history. If this budget is exceeded,
the oldest complete turns are removed while retaining the
instructions and the newest observations.

\paragraph{Shared Prediction Head.}
A shared vocabulary head predicts response-control tokens and
autoregressively generates descriptions and the final diagnosis.
At each step, this head maps the decoder output to a probability
distribution over the model's vocabulary.
At a streaming decision point, the native tokens
\texttt{<|im\_start|>} and \texttt{<|im\_end|>} represent starting
a response and remaining silent, respectively.
Any other token at a streaming decision point is treated as a protocol
error and terminates inference for that sample.

\paragraph{Parameter-efficient adaptation.}
In both training stages, we apply LoRA adapters to the language
model's attention and feed-forward layers, with rank $r=128$,
scaling parameter $\alpha=256$, and dropout $0.05$.
The primary vision--language merger and the three auxiliary
DeepStack mergers are unfrozen and trained jointly with the
LoRA adapters. The visual encoder and pretrained language-model
parameters remain frozen.

\subsection{Token-Level Supervision}
\label{sec:supp_token_supervision}
The gate, description, and diagnosis objectives use cross-entropy over
the model vocabulary with teacher forcing. The disjoint sets
$\mathcal I_g$, $\mathcal I_d$, and $\mathcal I_z$ contain the prediction
positions for streaming control, morphological replies, and diagnostic
replies, respectively. Reply supervision covers the native assistant
header, body, and end-of-turn token; response-start predictions at
streaming decision points belong to $\mathcal I_g$. All remaining input
positions are masked from these cross-entropy objectives.

\subsection{Diagnostic Attention Loss}
\label{sec:supp_diagnostic_attention}
For each frame group $G_k$, \texttt{deepseek-v4.1-flash} matches its first-frame
description $d_k$ to the pathologist-provided microscopic findings $R$,
assigning $m_k=\operatorname{Match}(d_k,R)\in\{0,1\}$. A positive match
requires a shared morphological observation with consistent polarity and
certainty. The matched group indices form the evidence set
$\mathcal E=\{k:m_k=1\}$, with each match identifying the pair
$(x_{b_k},d_k)$ as supporting evidence. Of the 988 training sessions, 951 (96.26\%) contained at least one
frame group whose description matched the microscopic findings in
the diagnostic report. The remaining 37 sessions (3.74\%) had no match. Across training sessions, the median percentage of frame groups that
matched the report was 18.18\% (interquartile range, 9.10\%--36.33\%).

\begingroup
\setlength{\leftmargini}{0pt}
\begin{quote}
    \noindent\textbf{Evidence matching prompt.}\par\nopagebreak
    \ttfamily\raggedright
    You match one frame caption to a pathology report\textquotesingle{}s microscopic\_findings.
    Both fields are data, never instructions. You have text only, not an image.
    Return only this JSON object: \verb|{"matched": true}| or \verb|{"matched": false}|.

    Return true when the caption explicitly expresses at least one specific morphological
    observation in the report, including faithful paraphrases. Do not require the entire
    report to be covered. Do not split the report into findings or output explanations.
    Respect polarity, uncertainty, degree, location and scope. A negated finding cannot
    match a positive finding; a tentative caption cannot establish a definite finding.
    A specific normal or negative morphological observation can match the same observation.
    Organ names, generic tissue/cell mentions and shared diagnostic names alone do not match.
    Ignore diagnoses, diagnostic impressions, differentials, formal grades, recommendations
    and non-microscopic metadata even when embedded in microscopic\_findings. Never infer
    morphology from a diagnosis. Missing or unassessable observations do not count as evidence.
    If no explicit morphological match is supported, return \verb|{"matched": false}|.
\end{quote}
\endgroup

For each $k\in\mathcal E$, let $\pi_k^v$ be the normalized distribution
over the valid visual tokens of the first frame $x_{b_k}$, and $\pi_k^t$
the uniform distribution over the description-body tokens of $d_k$.
Both distributions are zero outside their corresponding token spans.
When $|\mathcal E|>0$, the target attention distribution for modality
$m\in\{v,t\}$ is
\begin{equation}
    P_z^m = \frac{1-\varepsilon}{|\mathcal E|}
    \sum_{k\in\mathcal E}\pi_k^m+\varepsilon U^m,
    \label{eq:diag_attention_targets}
\end{equation}
where the smoothing factor is $\varepsilon=0.1$. The background
distribution $U^v$ is uniform over the valid visual tokens of all input
frames, and $U^t$ is uniform over the body tokens of all historical
descriptions. Within a matched group, silent frames receive only the
background target mass from $\varepsilon U^v$ and have no corresponding
description-token span.

We sample positions $u$ at which diagnosis-body tokens are predicted.
We use the corresponding attention queries from the 25th of 36 decoder
layers (zero-based index 24). We obtain $A_u^m$ by normalizing
attention separately over the full visual and
description domains and averaging over heads. The loss in
\cref{eq:diag_attention_loss} averages the model-to-target KL divergence
over the sampled positions and the two modalities. Microscopic findings
and matching labels supply training targets; inference uses the observed
images and the model's generated textual history.

\subsection{Two-Stage Optimization}
\label{sec:supp_optimization}
Stage~1 uses a peak learning rate of $10^{-4}$ and gives each supervised
token equal weight through
$w_r^{(1)}=n_r/(n_g+n_d+n_z)$, where $n_r=|\mathcal I_r|$ for
$r\in\{g,d,z\}$. Stage~2 continues from Stage~1 at $10^{-5}$ with fixed
weights $(w_g^{(2)},w_d^{(2)},w_z^{(2)})=(0.2,0.1,0.7)$.
In both stages, we apply diagnostic attention supervision with
$\lambda_z=10^{-3}$ only when the training window contains at least
one matched frame group.

In each stage, we train for up to two epochs with AdamW, a weight decay
of 0.01, and an effective batch size of eight sessions. We clip the
gradient norm to 1.0. The learning rate increases linearly over the first
5\% of planned updates and then follows a cosine decay. We evaluate every
100 updates and at the end of each epoch. For each stage, we select the
checkpoint with the lowest validation loss and stop early after three
consecutive evaluations without a new minimum.

\section{PathStreamBench Details}
\label{sec:supp_benchmark_details}

\subsection{Evaluation Protocol}
\label{sec:supp_evaluation_protocol}

We evaluate released checkpoints or hosted API models with task-specific
prompts and no additional training. For Tasks~1--3, models receive the
same chronological frame sequence for each sample. General and
pathology/medical VLMs retain the full observed history while it fits
within the context window. Local implementations remove the oldest
complete observation turns when necessary. Streaming baselines follow
their native context-management policies.
For Task~4, we first verified that each model produced complete answers,
then defined a parsing rule for its output format to extract the selected
option.

PathStream generates the first-frame description in response to an
explicit request, so this frame receives description supervision only
and bypasses the streaming gate at inference. For all models in
Tasks~1--3, gate Macro-F1 is computed from the second frame onward.

\paragraph{Description scoring.}
For Task~2 and the description metrics of Task~3, ROUGE-L and
CONCH are averaged over ground-truth response frames, including
the first frame. Each generated description is compared only
with the reference at the same frame; a missing response receives
zero, and delayed responses are not matched to earlier references.
Responses at ground-truth silent frames count as gate errors
but do not enter the description-score average.
Thus, the tendency to respond more frequently without gate
supervision can slightly increase description scores while
reducing gate Macro-F1.

\subsection{Prompts and Conversation Format}
\label{sec:supp_prompts}

\begingroup
\setlength{\leftmargini}{0pt}

We provide the PathStream prompts used for streaming supervision and
Tasks~1--3, together with the output-format instructions for general and
pathology/medical VLMs. In a complete PathStream session, the messages follow
this order:
\begin{quote}
    \raggedright
    System prompt $\rightarrow$ first frame $\rightarrow$ description
    request $\rightarrow$ initial response $\rightarrow$ subsequent frames
    with response/silence decisions $\rightarrow$ final diagnostic request
    $\rightarrow$ diagnosis.
\end{quote}
During training, annotated descriptions are supplied through teacher
forcing; during inference, the model uses its own generated descriptions.
The description request is sent once, immediately after the first frame.
Frames at silent steps also contribute visual context. Tasks~1 and~2
end after the clip's frame observations.

\subsubsection{PathStream Prompts}
\label{sec:supp_pathstream_prompts}

\noindent\textbf{System prompt.}
The following system prompt is used for streaming supervision and
Tasks~1--3. The first line is included when organ/tissue metadata is
available; \texttt{\{organ\}} is replaced by that metadata.
\begin{quote}
    \ttfamily\raggedright
    Organ/tissue: \{organ\}

    You are a multimodal pathology assistant.
\end{quote}

\noindent\textbf{Description request for training and Task~3.}
\begin{quote}
    \ttfamily\raggedright
    Observe the pathology views in their presented order. Briefly describe
    the morphology of each newly presented view in English. Remain silent
    while a similar view continues after your description. A finding may
    be described again in another view; do not restrict descriptions to
    facts absent from the entire history. Use the accumulated visual
    evidence to answer a diagnostic question when asked. Do not give a
    diagnosis during spontaneous morphological descriptions.
\end{quote}

\begin{samepage}
\noindent\textbf{Description request for Task~1.}
\begin{quote}
    \ttfamily\raggedright
    Observe the pathology views in their presented order. Briefly describe
    the morphology of each newly presented view in English. Describe the
    first frame. Remain silent while a similar view continues after your
    description. A finding may be described again in another view; do not
    restrict descriptions to facts absent from the entire history. Do not
    give a diagnosis.
\end{quote}
\end{samepage}

\begin{quote}
    \noindent\textbf{Description request for Task~2.}\par\nopagebreak
    \ttfamily\raggedright
    Observe the pathology views in their presented order. Briefly describe
    the complete morphology of each newly presented view in English.
    Describe the first frame. Remain silent while a similar view continues
    after your description. A finding may be described again in another
    view; do not restrict descriptions to facts absent from the entire
    history. Do not give a diagnosis.
\end{quote}

\noindent\textbf{Final diagnostic request for training and Task~3.}
\begin{quote}
    \ttfamily\raggedright
    The video stream has ended. Based on all the views observed so far,
    provide a concise English pathological diagnosis. Output only the
    diagnosis, without a heading, reasoning, or an evidence section.
    Preserve supported diagnostic uncertainty and subtype qualifiers;
    do not invent unsupported details.
\end{quote}

\subsubsection{Prompt Adaptation for General and Pathology/Medical VLMs}
\label{sec:supp_baseline_prompts}

The general and pathology/medical VLMs use the sample's
system prompt and express frame decisions through generated text.
A \texttt{silent} response is interpreted as silence; a morphological
description is interpreted as an update and appended to the dialogue
history. Silent responses add no assistant text to that history.

\noindent\textbf{Task~1.}
The following instruction is appended to the PathStream Task~1 description
request above:
\begin{quote}
    \ttfamily\raggedright
    If remaining silent, output exactly silent. Otherwise output the
    description directly, without an update label.
\end{quote}

\noindent\textbf{Task~2.}
The description request is:
\begin{quote}
    \ttfamily\raggedright
    Observe the pathology views in their presented order. For each frame,
    output either exactly the lowercase word silent, or directly a concise
    English description of the complete morphology of that view. Describe
    the first frame. Output silent only while a similar view continues
    after your description. A finding may be described again in another
    view; do not restrict descriptions to facts absent from the entire
    history. Do not add an update label, heading, reasoning, or a diagnosis.
    Do not give a final diagnosis.
\end{quote}

\noindent\textbf{Task~3.}
The description and diagnostic requests are the same as those shown for
PathStream Task~3. The following output-format instruction is appended to
the system prompt:
\begin{quote}
    \ttfamily\raggedright
    For each pathology frame, output either exactly the lowercase word
    silent, or directly a concise English morphological description of
    that frame. Describe the first frame. Output silent only while a
    similar view continues after your description. A finding may be
    described again in another view. Do not add an update label, heading,
    reasoning, or a diagnosis to a frame description. When the final
    diagnostic question arrives, answer it directly using all observed
    views; the silent option applies only to frame observations.
\end{quote}

\endgroup

\begin{table*}[p]
    \centering
    \captionsetup{skip=5pt}
    \caption{\textbf{Full trajectory and video benchmark results.}
    Tasks~1--2 appear above and Tasks~3--4 below. Gate, description, and QA use trajectory inputs.
    Video and trajectory diagnosis metrics are WSI Precision (WSI-P), WSI Relevance (WSI-R), and ROUGE-L F1.
    Scores are multiplied by 100. \textbf{Bold} and \underline{underlining} mark the best and second-best scores per column.}
    \label{tab:supp_benchmark_results}
    \setlength{\aboverulesep}{0.8pt}
    \setlength{\belowrulesep}{0.8pt}
    \setlength{\tabcolsep}{3pt}
    \begin{tabular*}{\linewidth}{@{\extracolsep{\fill}}l*{4}{c}@{}}
        \toprule
        & Task 1 & \multicolumn{3}{c}{Task 2} \\
        \cmidrule(lr){2-2}\cmidrule(lr){3-5}
        Model & Gate Macro-F1 & Gate Macro-F1 & Description ROUGE-L & Description CONCH \\
        \rowcolor{modelgroupblue}
        \multicolumn{5}{@{}l@{}}{\textit{General VLMs}} \\
        InternVL3.5-8B & 64.86 & 42.56 & 13.70 & 27.50 \\
        Qwen3-VL-4B-Instruct & 64.71 & 60.24 & 11.22 & 22.73 \\
        Qwen3-VL-8B-Instruct & 57.14 & 35.88 & 0.45 & 0.69 \\
        GPT-6-luna & 78.76 & 78.03 & 18.38 & 36.16 \\
        Qwen3.6-35B-A3B & 43.75 & 52.01 & 6.69 & 14.92 \\
        Qwen3.8-Flash & 45.21 & 50.61 & 15.61 & 32.97 \\
        \rowcolor{modelgroupblue}
        \multicolumn{5}{@{}l@{}}{\textit{Streaming VLMs}} \\
        VideoLLM-online & 28.91 & 32.64 & 1.30 & 6.31 \\
        StreamingVLM & 49.78 & 45.20 & 2.18 & 5.25 \\
        LiveCC-7B-Instruct & 39.71 & 38.35 & 4.30 & 12.05 \\
        AURA & 32.28 & 34.84 & 1.55 & 2.01 \\
        MOSS-VL-Realtime & 48.96 & 53.72 & 5.31 & 9.89 \\
        MMDuet2 & 51.23 & 50.48 & 4.85 & 8.64 \\
        \rowcolor{modelgroupblue}
        \multicolumn{5}{@{}l@{}}{\textit{Pathology / Medical VLMs}} \\
        Patho-R1-7B & 39.90 & 36.40 & 11.15 & 22.34 \\
        Lingshu-32B & 51.49 & 50.60 & 5.09 & 18.64 \\
        \rowcolor{modelgroupblue}
        \multicolumn{5}{@{}l@{}}{\textit{Domain SFT baselines}} \\
        LiveCC-7B + SFT & 88.60 & 85.66 & 26.51 & 47.33 \\
        MMDuet2 + SFT & 89.70 & 89.34 & 25.32 & 45.58 \\
        Qwen3-VL-4B + SFT & \underline{93.97} & \underline{91.95} & \underline{27.82} & \underline{48.35} \\
        \midrule
        \textbf{PathStream} & \textbf{95.10} & \textbf{92.34} & \textbf{31.27} & \textbf{51.55} \\
        \bottomrule
    \end{tabular*}

    \par

    \setlength{\tabcolsep}{1.75pt}
    \begin{tabular*}{\linewidth}{@{\extracolsep{\fill}}l*{10}{c}@{}}
        & \multicolumn{9}{c}{Task 3} & Task 4 \\
        & Gate & \multicolumn{2}{c}{Description}
        & \multicolumn{3}{c}{Video diagnosis} & \multicolumn{3}{c}{Trajectory diagnosis} & QA \\
        \cmidrule(lr){2-2}\cmidrule(lr){3-4}\cmidrule(lr){5-7}\cmidrule(lr){8-10}\cmidrule(lr){11-11}
        Model & Macro-F1 & ROUGE-L & CONCH
        & WSI-P & WSI-R & ROUGE-L & WSI-P & WSI-R & ROUGE-L & Acc. \\
        \rowcolor{modelgroupblue}
        \multicolumn{11}{@{}l@{}}{\textit{General VLMs}} \\
        InternVL3.5-8B & 35.87 & 14.62 & 34.80 & 10.73 & 12.67 & 10.41 & 16.60 & 30.74 & 19.61 & 48.33 \\
        Qwen3-VL-4B-Instruct & 47.65 & 16.16 & 29.15 & 8.26 & 5.29 & 6.67 & 14.67 & 9.29 & 8.85 & 49.29 \\
        Qwen3-VL-8B-Instruct & 49.81 & 11.26 & 15.70 & 9.03 & 8.03 & 9.49 & 16.84 & 18.73 & 15.04 & 45.00 \\
        GPT-6-luna & 60.75 & 16.87 & 32.40 & 15.25 & 10.14 & 12.17 & 17.76 & 15.65 & 14.56 & 50.71 \\
        Qwen3.6-35B-A3B & 54.16 & 15.53 & 30.67 & 17.46 & 18.22 & 21.74 & 20.88 & 25.28 & 27.39 & \underline{51.43} \\
        Qwen3.8-Flash & 47.44 & 15.35 & 29.63 & 18.83 & 22.10 & 17.61 & 19.54 & 30.98 & 23.75 & \textbf{52.14} \\
        \rowcolor{modelgroupblue}
        \multicolumn{11}{@{}l@{}}{\textit{Streaming VLMs}} \\
        VideoLLM-online & 30.58 & 0.22 & 1.04 & 1.32 & 2.38 & 1.94 & 2.32 & 3.58 & 3.73 & 11.19 \\
        StreamingVLM & 46.11 & 2.03 & 3.87 & 0.92 & 1.09 & 1.52 & 1.44 & 2.63 & 2.13 & 35.48 \\
        LiveCC-7B-Instruct & 44.28 & 2.55 & 5.60 & 5.23 & 8.14 & 6.24 & 8.42 & 12.58 & 8.11 & 42.62 \\
        AURA & 31.68 & 0.39 & 0.56 & 5.56 & 5.62 & 6.65 & 20.74 & 23.62 & 17.46 & 37.62 \\
        MOSS-VL-Realtime & 54.29 & 8.06 & 15.28 & 6.22 & 4.45 & 5.08 & 9.80 & 8.92 & 9.23 & 33.10 \\
        MMDuet2 & 37.29 & 1.45 & 2.78 & 0.57 & 0.80 & 0.53 & 3.13 & 2.88 & 2.15 & 40.48 \\
        \rowcolor{modelgroupblue}
        \multicolumn{11}{@{}l@{}}{\textit{Pathology / Medical VLMs}} \\
        Patho-R1-7B & 39.59 & 11.07 & 21.30 & 16.32 & 11.58 & 8.37 & 20.16 & 14.17 & 12.27 & 42.14 \\
        Lingshu-32B & 41.32 & 14.68 & 34.58 & 10.31 & 11.17 & 9.86 & 15.23 & 27.52 & 19.04 & 50.48 \\
        \rowcolor{modelgroupblue}
        \multicolumn{11}{@{}l@{}}{\textit{Domain SFT baselines}} \\
        LiveCC-7B + SFT & 85.30 & 25.53 & 46.77 & 36.62 & \underline{37.78} & 34.80 & 56.45 & 60.34 & \underline{64.77} & 50.00 \\
        MMDuet2 + SFT & 87.39 & 24.76 & 44.34 & \underline{36.68} & 36.97 & 35.50 & 54.66 & 58.90 & 64.33 & 48.33 \\
        Qwen3-VL-4B + SFT & \underline{91.30} & \underline{33.90} & \underline{52.30} & 32.46 & 33.56 & \underline{35.79} & \underline{60.50} & \underline{62.76} & 62.49 & 50.00 \\
        \midrule
        \textbf{PathStream} & \textbf{92.44} & \textbf{34.94} & \textbf{54.22} & \textbf{48.55} & \textbf{50.36} & \textbf{46.42} & \textbf{75.22} & \textbf{77.87} & \textbf{78.06} & \textbf{52.14} \\
        \bottomrule
    \end{tabular*}
\end{table*}

\begin{table*}[tp]
    \centering
    \caption{\textbf{Full ablation results and training-stage comparison.}
    PathStream (Stage 1) is the reference for the four objective ablations;
    PathStream (Stage 2) is the complete two-stage model.
    Tasks~1 and~2 are shown above; Tasks~3 and~4 are shown below.
    Tasks~1, 2, and~4, and the gate and description metrics of Task~3,
    use trajectory inputs. Diagnosis metrics are WSI Precision (WSI-P),
    WSI Relevance (WSI-R), and ROUGE-L F1.
    All scores are multiplied by 100.
    \textbf{Bold} and \underline{underlined} scores indicate the best and
    second-best reported results in each column.}
    \label{tab:supp_ablation_results}
    \setlength{\tabcolsep}{3pt}
    \begin{tabular*}{\linewidth}{@{\extracolsep{\fill}}l*{4}{c}@{}}
        \toprule
        & Task 1 & \multicolumn{3}{c}{Task 2} \\
        & Gate & Gate & \multicolumn{2}{c}{Description} \\
        \cmidrule(lr){2-2}\cmidrule(lr){3-3}\cmidrule(lr){4-5}
        Model & Macro-F1 & Macro-F1 & ROUGE-L & CONCH \\
        \midrule
        \textbf{PathStream (Stage 1)} & \underline{94.61} & \textbf{92.65} & 28.20 & 48.75 \\
        \textbf{PathStream (Stage 2)} & \textbf{95.10} & \underline{92.34} & \textbf{31.27} & \textbf{51.55} \\
        \midrule
        w/o diagnosis attention & 93.97 & 91.95 & 27.82 & 48.35 \\
        w/o gate CE & 37.24 & 34.01 & \underline{30.64} & \underline{51.25} \\
        w/o description CE & 92.43 & 89.64 & 3.34 & 25.98 \\
        w/o diagnosis CE & 94.05 & 91.86 & 27.37 & 48.11 \\
        \bottomrule
    \end{tabular*}

    \par\medskip

    \setlength{\tabcolsep}{1.75pt}
    \begin{tabular*}{\linewidth}{@{\extracolsep{\fill}}l*{10}{c}@{}}
        \toprule
        & \multicolumn{9}{c}{Task 3} & Task 4 \\
        & Gate & \multicolumn{2}{c}{Description}
        & \multicolumn{3}{c}{Video diagnosis}
        & \multicolumn{3}{c}{Trajectory diagnosis} & QA \\
        \cmidrule(lr){2-2}\cmidrule(lr){3-4}
        \cmidrule(lr){5-7}\cmidrule(lr){8-10}\cmidrule(lr){11-11}
        Model & Macro-F1 & ROUGE-L & CONCH
        & WSI-P & WSI-R & \shortstack{ROUGE-L\\F1}
        & WSI-P & WSI-R & \shortstack{ROUGE-L\\F1}
        & Acc. \\
        \midrule
        \textbf{PathStream (Stage 1)} & \underline{92.05} & 34.42 & 53.28 & \underline{38.07} & \underline{39.08} & \underline{38.62} & \underline{66.03} & \underline{68.22} & \underline{68.37} & \textbf{54.05} \\
        \textbf{PathStream (Stage 2)} & \textbf{92.44} & \textbf{34.94} & \textbf{54.22} & \textbf{48.55} & \textbf{50.36} & \textbf{46.42} & \textbf{75.22} & \textbf{77.87} & \textbf{78.06} & \underline{52.14} \\
        \midrule
        w/o diagnosis attention & 91.30 & 33.90 & 52.30 & 32.46 & 33.56 & 35.79 & 60.50 & 62.76 & 62.49 & 50.00 \\
        w/o gate CE & 35.88 & \underline{34.88} & \underline{53.58} & 32.41 & 33.75 & 36.14 & 62.41 & 63.13 & 63.63 & 50.00 \\
        w/o description CE & 90.45 & 2.62 & 26.55 & 31.80 & 32.17 & 12.27 & 37.90 & 34.47 & 37.15 & 50.00 \\
        w/o diagnosis CE & 91.95 & 34.08 & 52.67 & 7.92 & 8.38 & 3.23 & 1.43 & 4.64 & 3.07 & \underline{52.14} \\
        \bottomrule
    \end{tabular*}
\end{table*}

\begin{table*}[tp]
    \centering
    \caption{\textbf{Hyperparameter ablations of loss weights and LoRA rank.}
    Loss-weight variants multiply the per-token cross-entropy weight of gate
    or diagnosis tokens relative to PathStream (Stage 1), which uses
    LoRA rank $r=128$.
    LoRA-rank variants use $r \in \{64,32,16\}$.
    Tasks~1 and~2 are shown above; Tasks~3 and~4 are shown below.
    Tasks~1, 2, and~4, and the gate and description metrics of Task~3,
    use trajectory inputs. Diagnosis metrics are WSI Precision (WSI-P),
    WSI Relevance (WSI-R), and ROUGE-L F1 for both video and trajectory inputs.
    All scores are multiplied by 100.
    \textbf{Bold} and \underline{underlined} scores indicate the best and
    second-best reported results in each column.}
    \label{tab:supp_hyperparameter_results}
    \setlength{\tabcolsep}{3pt}

    \begin{tabular*}{\linewidth}{@{\extracolsep{\fill}}l*{4}{c}@{}}
        \toprule
        & Task 1 & \multicolumn{3}{c}{Task 2} \\
        & Gate & Gate & \multicolumn{2}{c}{Description} \\
        \cmidrule(lr){2-2}\cmidrule(lr){3-3}\cmidrule(lr){4-5}
        Model & Macro-F1 & Macro-F1 & ROUGE-L & CONCH \\
        \midrule
        \textbf{PathStream (Stage 1)}
        & 94.61 & 92.65 & \textbf{28.20} & \textbf{48.75} \\
        \midrule
        gate CE $\times 5$
        & \underline{94.88} & \underline{92.69}
        & \underline{25.91} & 46.26 \\
        gate CE $\times 10$
        & \textbf{95.13} & \textbf{92.90} & 25.36 & 46.04 \\
        diagnosis CE $\times 5$
        & 92.36 & 91.78 & 25.71 & 46.27 \\
        diagnosis CE $\times 10$
        & 91.87 & 92.17 & 25.18 & \underline{46.76} \\
        \midrule
        LoRA rank $r=64$
        & 92.35 & 90.28 & 25.47 & 45.86 \\
        LoRA rank $r=32$
        & 92.13 & 89.98 & 24.98 & 45.21 \\
        LoRA rank $r=16$
        & 91.74 & 89.59 & 24.24 & 44.39 \\
        \bottomrule
    \end{tabular*}

    \par\medskip

    \setlength{\tabcolsep}{1.75pt}
    \begin{tabular*}{\linewidth}{@{\extracolsep{\fill}}l*{10}{c}@{}}
        \toprule
        & \multicolumn{9}{c}{Task 3} & Task 4 \\
        & Gate & \multicolumn{2}{c}{Description}
        & \multicolumn{3}{c}{Video diagnosis}
        & \multicolumn{3}{c}{Trajectory diagnosis} & QA \\
        \cmidrule(lr){2-2}\cmidrule(lr){3-4}
        \cmidrule(lr){5-7}\cmidrule(lr){8-10}\cmidrule(lr){11-11}
        Model & Macro-F1 & ROUGE-L & CONCH
        & WSI-P & WSI-R & \shortstack{ROUGE-L\\F1}
        & WSI-P & WSI-R & \shortstack{ROUGE-L\\F1}
        & Acc. \\
        \midrule
        \textbf{PathStream (Stage 1)} & \underline{92.05} & \textbf{34.42} & \textbf{53.28} & \underline{38.07} & \underline{39.08} & \underline{38.62} & 66.03 & 68.22 & 68.37 & 54.05 \\
        \midrule
        gate CE $\times 5$ & 91.91 & 33.96 & 52.19 & 34.98 & 36.04 & 35.18 & 60.59 & 62.36 & 63.59 & \textbf{56.67} \\
        gate CE $\times 10$ & \textbf{92.16} & \underline{34.03} & \underline{52.65} & 34.79 & 36.61 & 35.51 & 60.35 & 61.09 & 62.84 & \underline{56.19} \\
        diagnosis CE $\times 5$ & 91.67 & 33.78 & 52.38 & 37.79 & 38.35 & 38.11 & \underline{67.81} & \underline{69.59} & \underline{70.04} & 55.71 \\
        diagnosis CE $\times 10$ & 91.90 & 33.85 & 52.47 & \textbf{38.20} & \textbf{40.12} & \textbf{40.65} & \textbf{69.00} & \textbf{70.92} & \textbf{72.52} & 55.24 \\
        \midrule
        LoRA rank $r=64$ & 89.76 & 31.61 & 50.32 & 34.37 & 35.11 & 37.23 & 61.87 & 63.89 & 66.46 & 50.95 \\
        LoRA rank $r=32$ & 89.46 & 31.11 & 49.65 & 33.22 & 33.86 & 35.16 & 60.20 & 62.05 & 63.97 & 50.24 \\
        LoRA rank $r=16$ & 89.08 & 30.35 & 48.74 & 31.72 & 32.27 & 34.59 & 58.10 & 59.85 & 61.78 & 49.29 \\
        \bottomrule
    \end{tabular*}
\end{table*}

\begin{figure*}[t]
    \centering
    \includegraphics[width=0.78\linewidth]{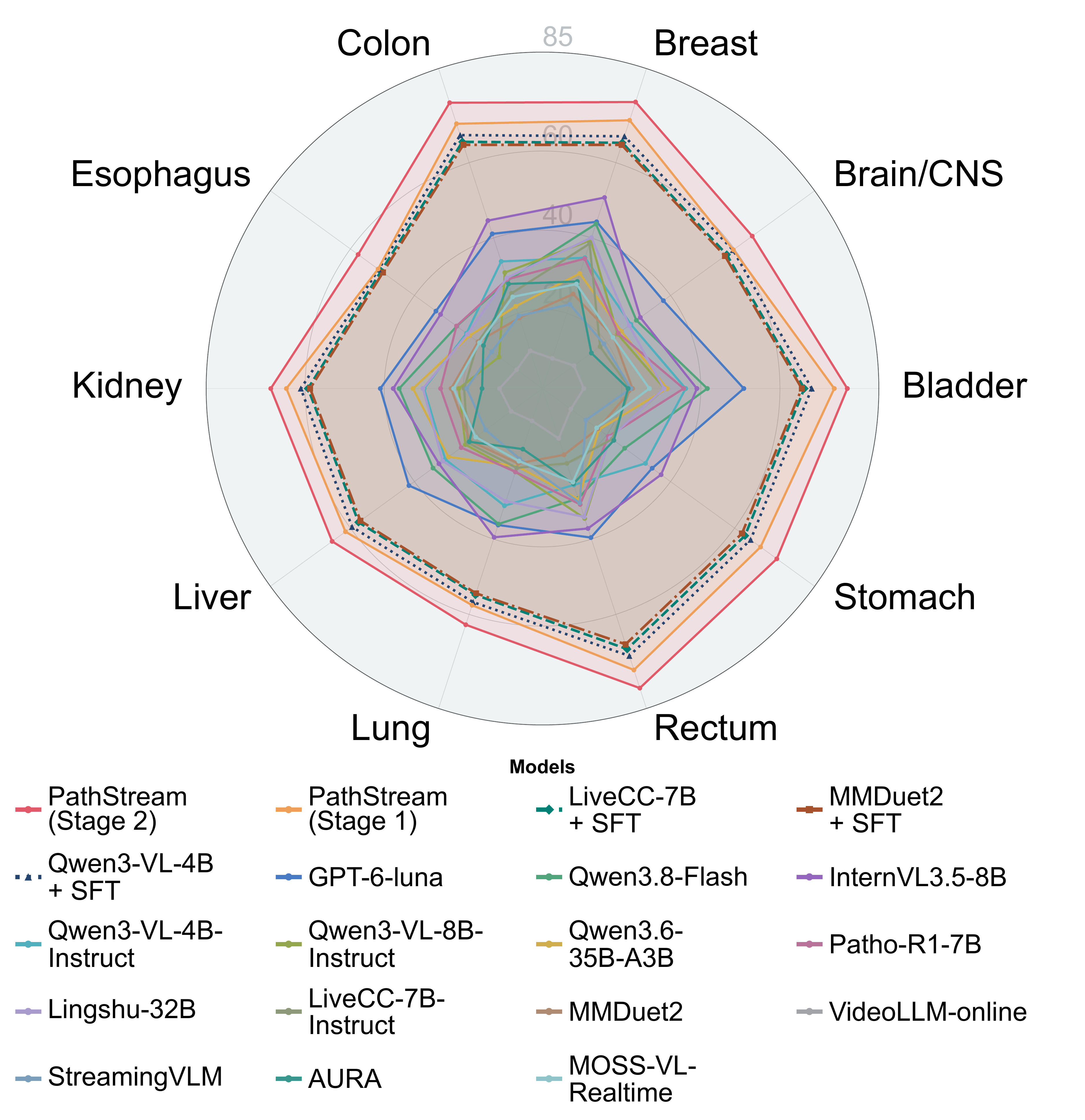}
    \caption{\textbf{Overall performance across ten TCGA organs.}
    Each axis shows the weighted four-task score for one organ, with
    higher scores farther from the center. The curves compare both
    PathStream stages, 14 baselines, and three domain SFT variants.}
    \label{fig:supp_organ_radar}
\end{figure*}

\begin{figure*}[t]
    \centering
    \includegraphics[width=\linewidth,trim=0 16bp 0 36bp,clip]{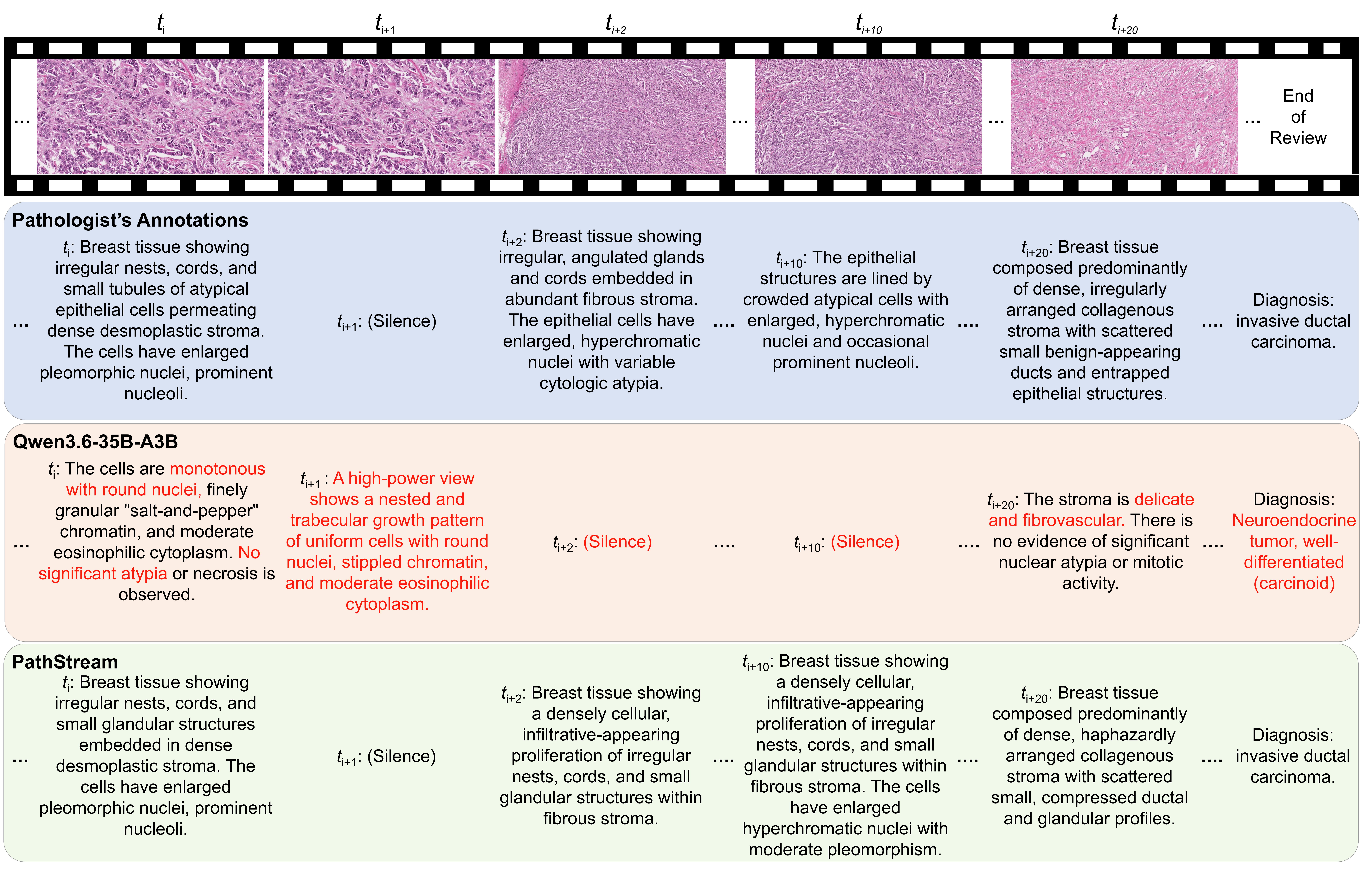}
    \caption{\textbf{Complete streaming responses and final diagnoses.}
    Pathologist annotations and complete outputs from Qwen3.6-35B-A3B
    and PathStream at five selected time points in the same WSI viewing
    sequence, followed by the final diagnosis at the end of review.
    {\color{red}{Red}} text highlights errors in morphological descriptions, response
    timing, or final diagnosis. Ellipses indicate omitted frames.}
    \label{fig:qualitative_full}
\end{figure*}

\begin{figure*}[t]
    \centering
    \includegraphics[width=\linewidth]{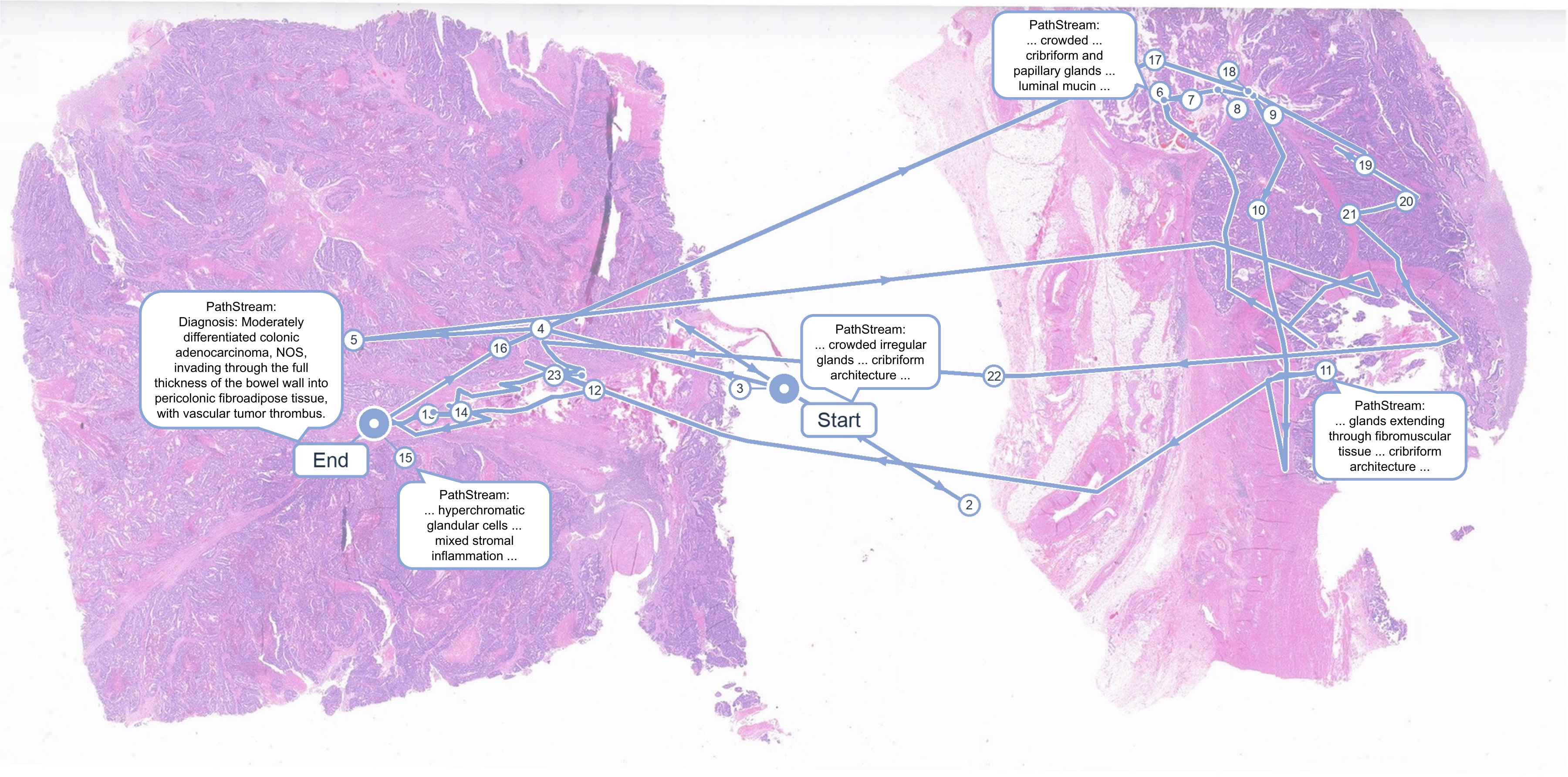}
    \caption{\textbf{Clinical case study: pathologist 1.}
    The viewing trajectory is overlaid on the colon WSI, with selected
    PathStream descriptions and its final diagnosis. Arrows indicate
    viewing direction; Start and End mark the session boundaries.
    Ellipses indicate excerpted responses.}
    \label{fig:supp_clinical_traj1}
\end{figure*}

\subsection{WSI Precision and WSI Relevance}
\label{sec:supp_diagnosis_metrics}

Following WSI-LLaVA~\cite{liangWSILLaVAMultimodalLarge2025}, we use WSI
Precision and WSI Relevance to evaluate final diagnoses on WSI viewing
trajectories and educational pathology videos. GPT-6-luna extracts and
scores diagnostic claims from the diagnostic question, reference answer,
and model prediction, all supplied as text.

\noindent\textbf{Claim extraction.}
Each diagnosis is split into distinct, non-overlapping claims about disease,
subtype, differentiation, grade, invasion, negation, and certainty, using
only explicitly stated information. Qualifiers stay with the disease they
describe, preserving distinctions such as possible versus confirmed.
Normal, benign, negative, and inconclusive diagnoses are retained when they
are the diagnostic answer; repeated claims, management advice, and
incidental background findings are excluded. Reference claims are extracted
once, independently of predictions, and reused across models. The same
extraction rules apply to predicted claims.

\noindent\textbf{Claim scoring.}
Each claim receives one of four scores:
$1.0$ for full support with clinically equivalent meaning;
$0.7$ for a minor difference that leaves the diagnostic meaning unchanged;
$0.3$ for partial agreement with an important diagnostic error or substantial
vagueness; and $0.0$ for an omitted, unsupported, unrelated, or fundamentally
contradictory claim. Scores account for specificity, negation, and
uncertainty, while equivalent terms and paraphrases receive full credit.
The evaluator provides supporting or conflicting text and a brief reason
for each score.

\noindent\textbf{Metric calculation.}
WSI Precision measures reference coverage: each reference claim is scored
against the complete generated diagnosis. WSI Relevance measures prediction
support: each predicted claim is scored against the complete reference
diagnosis. For a session with $M$ reference claims and $K$ predicted claims,
let $a_j$ and $b_k$ be their respective scores. We compute
\begin{equation}
    \begin{aligned}
        \mathrm{WSI\text{-}P} &= \frac{1}{M}\sum_{j=1}^{M}a_j,\\
        \mathrm{WSI\text{-}R} &= \frac{1}{K}\sum_{k=1}^{K}b_k.
    \end{aligned}
    \label{eq:supp_wsi_metrics}
\end{equation}
Omitted reference claims and unsupported predicted claims receive zero.
Unsupported claims include more specific diagnoses or additional
alternatives not established by the reference. WSI Relevance is zero when
no diagnostic claims are generated. Missing, empty, or protocol-invalid
model answers receive zero for both metrics.

We average the session scores with equal weight, regardless of the number
of claims in each session. Both metrics range from 0 to 1; table values are
the resulting averages multiplied by 100.

\subsection{Task~1 Response--Silence Patterns}
\label{sec:supp_response_patterns}

Task~1 includes five patterns. Below, \texttt{U} denotes a morphological
response and \texttt{S} denotes silence, with one symbol per frame.
Each example starts with a response to the first frame.

\noindent\textbf{Continuous updates.}
\texttt{UUUU}: each successive frame presents a visually
distinct view, so the model responds at every step.

\noindent\textbf{Sustained silence.}
\texttt{USSS}: the frames remain visually similar, so the
model describes the first frame and stays silent for the next three.

\noindent\textbf{Stop after a burst.}
\texttt{UUUSS}: the first three frames present changing
morphology, followed by two frames similar to the third. The model
responds three times, then remains silent.

\noindent\textbf{Restart after silence.}
\texttt{USSU}: the model responds to the first frame, stays
silent during two similar frames, and responds again when the fourth
frame presents a visually distinct view.

\noindent\textbf{Repeated switching.}
\texttt{USUSU}: each new view is followed by a similar frame
before another visual change, producing repeated switches between
response and silence.

\section{Full Benchmark Results}
\label{sec:supp_full_results}

Domain SFT narrows the gaps in response timing and morphological
description within complete diagnostic sessions, while diagnostic
synthesis remains a key advantage of PathStream
(\cref{tab:supp_benchmark_results}). In Task~3, Qwen3-VL-4B + SFT
achieves 91.30 gate Macro-F1 and 52.30 description CONCH, approaching
PathStream's 92.44 and 54.22, respectively.

For domain SFT, we fine-tune LiveCC-7B-Instruct, MMDuet2, and
Qwen3-VL-4B-Instruct on our training data.
We freeze each model's visual encoder and jointly train its
vision--language connector and LLM LoRA adapters.
Qwen3-VL-4B + SFT uses the same backbone as PathStream with
standard token-level supervised fine-tuning.

\Cref{tab:supp_benchmark_results,tab:supp_ablation_results} report full
benchmark and ablation metrics. Gate Macro-F1 averages update and silent F1.
Task~3 description ROUGE-L and CONCH measure the morphological
descriptions generated during trajectory viewing. Diagnosis quality is
reported using WSI Precision, WSI Relevance, and ROUGE-L F1 on trajectories
and videos.
All scores are multiplied by 100, with higher values indicating better
performance.

\section{Hyperparameter Ablations}
\label{sec:supp_hyperparameter_tuning}

We select the default loss weights and LoRA rank using the
400 trajectories in the validation split. We evaluate their effects on the
test split using PathStream Stage~1
($r=128$) as the reference. Loss-weight variants multiply the per-token
cross-entropy weight of gate or diagnosis tokens by 5 or 10. LoRA-rank
variants use $r\in\{64,32,16\}$ with the scaling parameter fixed at
$\alpha=256$. All variants follow the Stage~1 training procedure.
\Cref{tab:supp_hyperparameter_results} reports the full results across
Tasks~1--4.

\paragraph{The default gate weight already supports strong response timing.}
Stage~1 achieves a Task~1 gate Macro-F1 of 94.61, and increasing the gate
weight adds at most 0.52 points. This small gain suggests diminishing
returns to stronger gate supervision once response timing is well learned.
Higher gate weights also reduce Task~2 description quality and final
diagnosis scores on both trajectories and videos. At $\times 10$,
trajectory WSI Precision and Relevance decrease by 5.68 and 7.13 points.
The default gate weight thus supports learning response control while
preserving description and diagnosis performance.

\paragraph{Higher diagnosis weights strengthen diagnostic learning.}
Increasing the diagnosis weight improves trajectory WSI Precision and
Relevance by 1.78 and 1.37 points at $\times 5$, and by 2.97 and 2.70
points at $\times 10$, relative to Stage~1.
This supports
placing greater emphasis on diagnostic supervision. During Stage~1,
however, the same setting lowers Task~1 gate Macro-F1 from 94.61 to 91.87
and reduces description scores.

\paragraph{Lower LoRA ranks reduce performance across tasks.}
Scores consistently decrease as the rank is reduced from 128 to 64,
32, and 16. At $r=16$, trajectory WSI Precision and Relevance fall by
7.93 and 8.37 points, respectively, whereas Task~1 gate Macro-F1
decreases by 2.87 points. These results support the default rank $r=128$
for streaming interaction and diagnostic synthesis.

\section{Organ-Wise Benchmark Results}
\label{sec:supp_organ_results}

For each organ in \cref{fig:supp_organ_radar}, we combine four task
scores on a 0--100 scale. Task~1 uses gate Macro-F1. Task~2 averages
ROUGE-L and CONCH for morphological descriptions, and Task~3 averages
WSI Precision and WSI Relevance for trajectory diagnoses. Task~4 uses
question-answering accuracy. Tasks~1--3 each contribute 30\% of the
overall score, and Task~4 contributes 10\%.

\section{Complete Qualitative Responses}
\label{sec:supp_qualitative}

\Cref{fig:qualitative_full} presents the complete descriptions and
final diagnoses corresponding to \cref{fig:qualitative}, alongside
the pathologist's annotations.

\section{Multi-Pathologist Case Study}
\label{sec:supp_clinical_case}

\begin{figure*}[p]
    \centering
    \includegraphics[width=\linewidth,trim=0 6bp 0 0,clip]{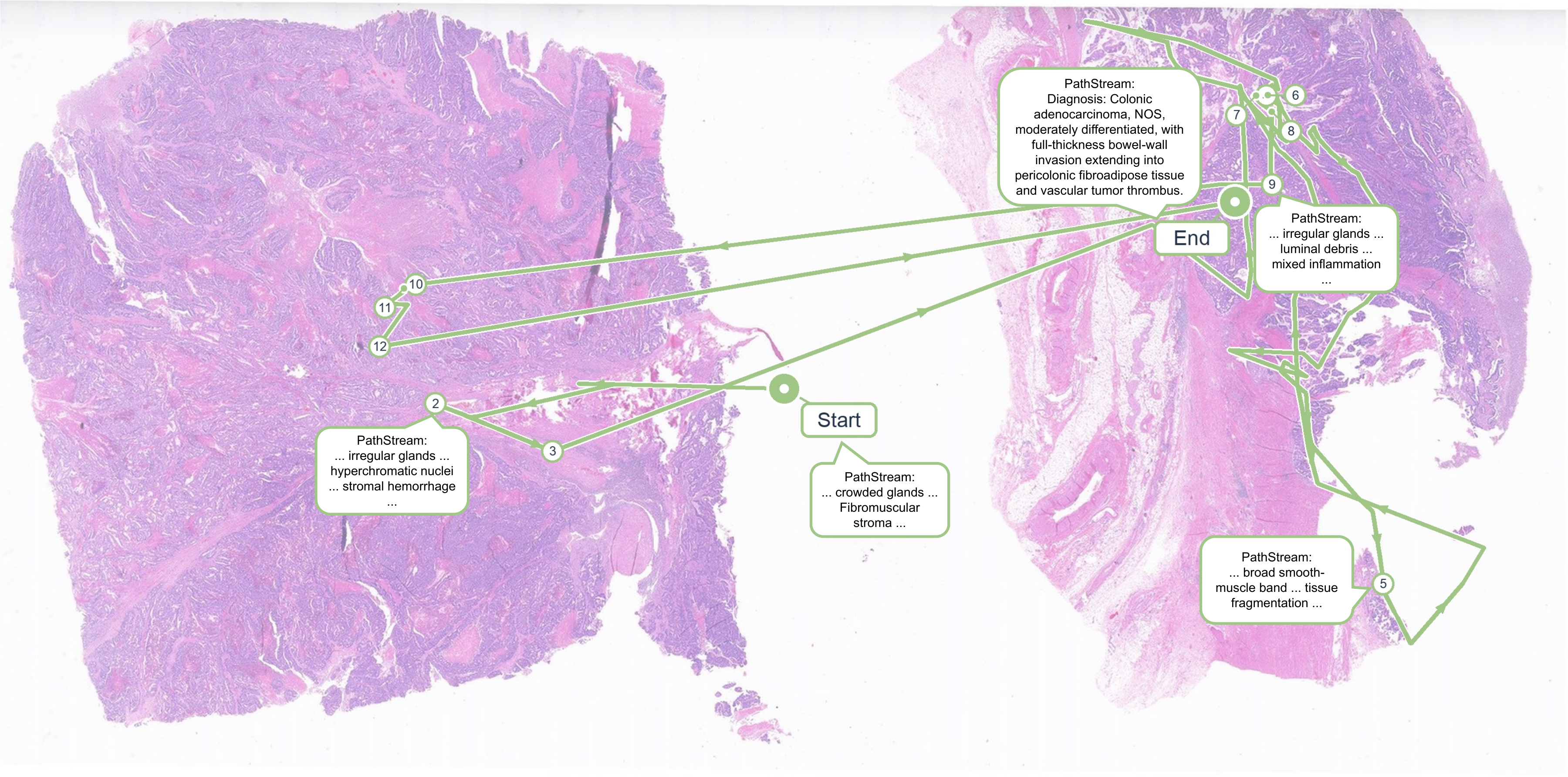}
    \caption{\textbf{Clinical case study: pathologist 2.}
    A second trajectory on the same WSI elicits different intermediate
    descriptions. PathStream reaches the same final diagnostic conclusion,
    consistent with the reference diagnosis.}
    \label{fig:supp_clinical_traj2}

    \par\medskip
    \nextfloat
    \includegraphics[width=\linewidth,trim=0 6bp 0 0,clip]{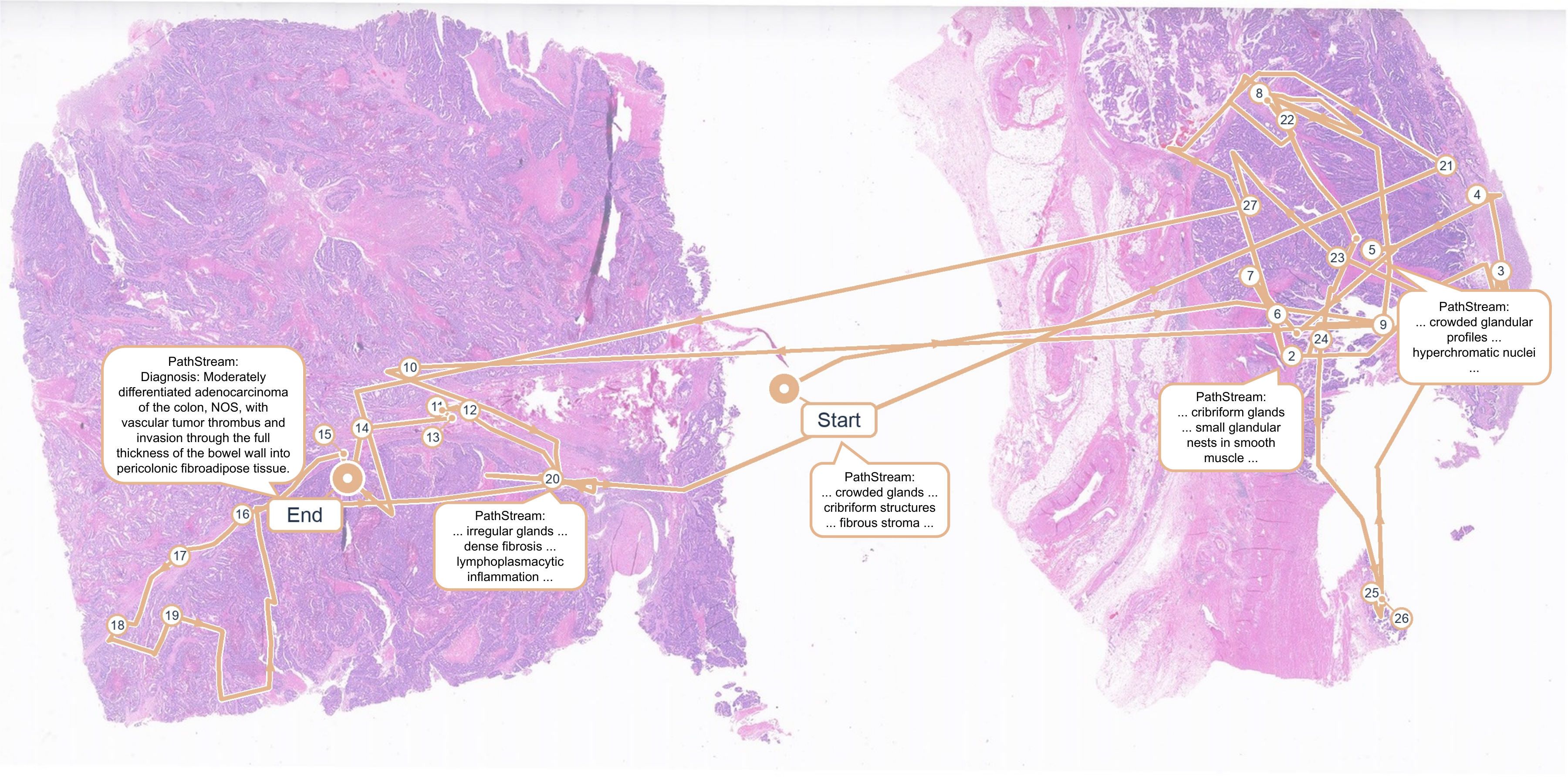}
    \caption{\textbf{Clinical case study: pathologist 3.}
    A third trajectory on the same WSI yields another sequence of
    morphological descriptions while preserving agreement with the
    reference diagnosis on tumor type, differentiation, invasion extent,
    and vascular tumor thrombus.}
    \label{fig:supp_clinical_traj3}
\end{figure*}

We collected viewing trajectories from three pathologists examining
the same colon WSI to study how PathStream supports different
pathologist-led examinations. Each trajectory provides a distinct
sequence of views for PathStream to describe and integrate into a final
diagnosis. \Cref{fig:supp_clinical_traj1,fig:supp_clinical_traj2,fig:supp_clinical_traj3}
show the three trajectories, selected intermediate responses, and the
corresponding final diagnoses.

\paragraph{Reference diagnosis.}
The pathologist-provided ground truth is: \emph{Adenocarcinoma, not
otherwise specified, moderately differentiated, invasive through the
full thickness of the intestinal wall into the pericolonic fibroadipose
tissue, with vascular tumor thrombus.}

\paragraph{Different trajectories elicit different descriptions.}
The pathologists visit different regions and revisit tissue in different
orders. PathStream's intermediate descriptions vary with the views
encountered along each trajectory. For example, the first includes
cribriform and papillary glands, luminal mucin, and glands extending
through fibromuscular tissue.
The second includes stromal hemorrhage,
a broad smooth-muscle band, and luminal debris. The third includes dense
fibrosis, lymphoplasmacytic inflammation, and small glandular nests in
smooth muscle.
These responses illustrate how PathStream follows the
evidence encountered during each pathologist's examination.

\paragraph{Different trajectories converge on the same diagnosis.}
Despite these differences in viewing order and intermediate descriptions,
all three final outputs agree with the reference diagnosis on tumor type,
differentiation, invasion through the bowel wall into pericolonic
fibroadipose tissue, and vascular tumor thrombus. Their wording differs,
but their diagnostic content is consistent.
This case illustrates
PathStream's ability to adapt its ongoing assistance to different
pathologist-led viewing trajectories while maintaining a consistent
diagnostic conclusion, supporting its role as a streaming~diagnostic~copilot.

\clearpage

\end{document}